\documentclass[a4paper,fleqn]{cas-dc}

\usepackage{float}
\usepackage[numbers]{natbib}
\usepackage{graphicx}
\usepackage[utf8]{inputenc}
\usepackage[russian, english]{babel}
\usepackage{lineno}
\usepackage[singlelinecheck=false]{caption}
\usepackage{xcolor}
\setpagewiselinenumbers
\let\printorcid\relax

\def\tsc#1{\csdef{#1}{\textsc{\lowercase{#1}}\xspace}}
\tsc{WGM}
\tsc{QE}
\tsc{EP}
\tsc{PMS}
\tsc{BEC}
\tsc{DE}

\begin{document}
\let\WriteBookmarks\relax
\def\floatpagepagefraction{1}
\def\textpagefraction{.001}
\shorttitle{Detecting Visual Design Principles in Art and Architecture through Deep
Convolutional Neural Networks}
\shortauthors{}

\title{Detecting Visual Design Principles in Art and Architecture through Deep Convolutional Neural Networks}

\author[1]{Gözdenur Demir}[
     type=author,
]
 \ead{demirg@itu.edu.tr}
\address[1]{Istanbul Technical University, Faculty of Architecture, Istanbul, 34437, Turkey}

\author[1]{Aslı Çekmiş}[
     type=author,
]
 \ead{cekmis@itu.edu.tr}
\cormark[1]

\author[2]{Vahit Bugra Yesilkaynak}[
    type=author,
   ]
 \ead{yesilkaynak15@itu.edu.tr}
\address[2]{Istanbul Technical University, Faculty of Computer and Informatics Engineering, Istanbul, 34469, Turkey}

\author[2]{Gozde Unal}[
     type=author,
]
 \ead{gozde.unal@itu.edu.tr}

\cortext[0]{This work is to be published at the Journal "Automation in Construction". https://doi.org/10.1016/j.autcon.2021.103826 }

\begin{abstract}
Visual design is associated with the use of some basic design elements and principles. Those are applied by the designers in the various disciplines for aesthetic purposes, relying on an intuitive and subjective process. Thus, numerical analysis of design visuals and disclosure of the aesthetic value embedded in them are considered as hard. However, it has become possible with emerging artificial intelligence technologies. This research aims at a neural network model, which recognizes and classifies the design principles over different domains. The domains include artwork produced since the late 20th century; professional photos; and facade pictures of contemporary buildings. The data collection and curation processes, including the production of computationally-based synthetic dataset, is genuine. The proposed model learns from the knowledge of myriads of original designs, by capturing the underlying shared patterns. It is expected to consolidate design processes by providing an aesthetic evaluation of the visual compositions with objectivity.
\end{abstract}




\maketitle

\section{Introduction}

The procedure that transforms a brief or requirement into a completed final product or solution is called design \cite{ambrose2009basics}. Visual design is the tangible representation of the objectives of a product and is concerned with the ‘look’, ‘method’ and ‘style’ in which the information is presented. The language of visual design is creatively employed by artists, designers and architects \cite{kress1996reading}. \

Although not transparent or universally understood, visual design depends on various design elements and principles. These are applied in all art disciplines \cite{fichner2011foundations}, such as painting, drawing, sculpture and photography \cite{hirsch2012light}; in craft disciplines, such as ceramics, textiles and glass; in applied design fields, such as graphical design \cite{peterson2003using,landa2010graphic,arntson2011graphic,white2011elements} and industrial design; and in environmental design disciplines, such as architecture \cite{ching2014architecture}, interior design, landscape design \cite{bell2019elements} and urban planning. Design elements and principles are used for two-dimensional surfaces (as in graphical design), or three-dimensional forms (as in architecture). The way in which the design elements are visually organized could be defined in various ways, such as a composition \cite{krause2004design}, a structure \cite{peterson2003using}, a visual pattern \cite{lauer2011design}, and a visual language system \cite{puhalla2011design}. The logical procedures that use design elements (e.g. points, lines, planes and attributes like color, size and shape) to establish a perceptual framework for visual processing \cite{puhalla2011design} are known as visual design principles (VDP). They include harmony, unity, balance, rhythm, emphasis, proportion, contrast, movement, and repetition \cite{landa2010graphic}. Similarly, the perceptual organization of the small parts into wholes is defined by a series of rules –a set of Gestalt principles– such as similarity, proximity, continuation, closure, and figure and ground \cite{arntson2011graphic}. Despite the terms ‘Gestalt principles’ and ‘VDP’ being used interchangeably \cite{lin2013development}, this research will use concepts developed from VDP. \

Peterson \cite{peterson2003using} states that VDP are the bylaws of a proper structure, which are formally taught in visual design disciplines. Regarding their use, Watzman \cite{watzman2003visual} states that “good visual interaction and experience design bridge many worlds: that of visual design, information presentation, and usability with that of aesthetics” (p. 266). So, visual aesthetics is an important criterion for designers when attempting to convey information through visual design composition by means of the proper employment of various VDP. \

It is generally believed that the designers use VDP instinctively in the unique composition of a design object. Despite the individuality pertaining to an artwork, Arnheim \cite{arnheim1974art} states that “... a well-organized line figure imposes itself upon all observers as basically the same shape... one could expect the same, at least in principle, with respect to people looking at the works of art. This trust in the objective validity of the artistic statement supplied a badly needed antidote to the nightmare of unbounded subjectivism and relativism” (p. 6). This leads the way to more objective interpretations of art. The analysis of visual compositions is necessary for the sake of design explanation and exploration and can convey an insight into rethinking the ‘recipe’ of design. Arnheim \cite{arnheim1974art} also points out that “vision is not a mechanical recording of elements but rather the apprehension of significant structural patterns” (p. 6). Following this notion, this research asserts that the common rules of ‘visual things’ –the VDP– could be identified with underlying computational patterns. \

The way humans process and interpret visual information from two-dimensional surfaces has long been analyzed by researchers, including Gestalt psychologists \cite{arntson2011graphic}. Suggested evaluative computational methods for analyzing Gestalt principles and VDP show the necessity of a well-thought-of approach that is capable of capturing the principles conceptually \cite{lin2013development}. However, the complexity of visual design compositions and the relations of design elements with each other makes this a challenge. In recent years ‘deep convolutional neural networks’ (abbreviated as CNNs throughout the paper), from the field of computer vision in artificial intelligence (AI), have produced simple yet effective models of the visual system \cite{lindsay2020convolutional}. They are highly appreciated for their success and progress in image classification tasks and for surpassing human-level performance on several single-label image classification tasks \cite{rawat2017deep}. CNNs have a great deal of potential for the analysis of visual composition and their use has already been explored for recognition of Gestalt principles \cite{ehrensperger2019evaluating}, but not of VDP. Such a model would provide fast, objective and numerical deciphering of visual compositions through VDP. \

We employ an AI model that detects the VDP within a visual composition mathematically and reveals the qualitative and quantitative aspects of the data. So, it is possible to observe and test the existing visual rules and hypotheses presented within the visual design field, for instance, relations and co-occurrences. Specifically, we adapted a recent neural network model, EfficientNet \cite{tan2019efficientnet}, and optimized it for VDP classification. The VDP lead to different patterns of content, frequency and types of usage in data from different domains. Our model, therefore, analyses data from different domains of visual design: photography (PHT), abstract art (ART) and architecture (ARC). A well-prepared dataset in all relevant domains is a necessary asset in AI, and consequently the data curation process (including tasks such as searching for, collecting, augmenting and pruning the data) is of great importance. \

This article is structured so that readers are introduced to relevant prior research in the next section. Section 3 describes the quantification of the VDP, which will build the logical structure of learning algorithms –depending on the various forms of expression of principles. Section 4 presents the work of collecting and analyzing data, followed by the explanation of the CNN model in Section 5. Section 6 discusses the model performance and, finally, Section 7 concludes the paper by highlighting the limitations and future extensions of the research. \

\section{Using computers for visual analysis in art and architecture}
\subsection{AI-based studies and motivations in art}

Nowadays, increasing numbers of studies are carried out in various subjects that involve using AI with paintings and photographs from art history. These studies include style classification and style recognition –for identifying an artist’s painting \cite{blessing2010using}; photograph and painting discrimination –for exploring the ability of complexity-based metrics \cite{carballal2018distinguishing}; visual saliency detection –for modeling the way human eye sees and understands visual art through detection and analysis of regions of interest \cite{condorovici2011saliency}; emotion recognition –for reading positive or negative emotions in artworks \cite{yanulevskaya2012eye}; visual aesthetic analysis –for getting insight into the attributes associated with aesthetically pleasing images \cite{ciesielski2013finding}; visual complexity analysis –for making accurate predictions about humans’ impression of the visual complexity of objects, scenes or designs \cite{machado2015computerized} and for defining factors that affect visual complexity perception \cite{guo2018assessment}; and lastly art generation with creative characteristics –for human-level forming of original ideas through exploration and discovery \cite{elgammal2017can}. To sum up, these studies show that AI is currently widely applied in the field of art, and research areas are gradually expanding. Direct quantification and detection of VDP with AI has not yet been performed, except in a study by Cetinic et al. \cite{cetinic2020learning} which aims to quantify stylistic properties and to predict the values of Wölfflin’s visual principles. \

\subsection{AI-based studies in architecture}

With recent developments in image processing and machine learning (ML) and their experimental applications in various fields including building design \cite{loyola2018big}, built environment data can now be analyzed computationally. Some of the architectural data used for conducting specific learning tasks includes street view images \cite{liu2017machine}, aerial images \cite{liu2014local}, technical drawings \cite{tome2015space,huang2018architectural}, design sketches \cite{karimi2020creative} and facades. This research focuses on the ‘facade,’ as the 2D (front) view of a building, which is given an artistic appearance. \

\subsubsection{Visual analysis of facades}

Previous studies have explored different approaches and methods for the visual analysis of facades. Researchers focused on concepts, such as complexity \cite{bovill1996fractal,salingaros1997life,imamoglu2000complexity,stamps2003advances}, variety \cite{cooper2008fractal} and visual aesthetics \cite{nasar1994urban,gifford2000decoding}. However, studies of design elements and principles in architectural facades are relatively limited. Hasse and Weber \cite{hasse2012eye} ran an experiment on the evaluation of balance and looked for a correlation between beauty and balance with subject evaluations and eye movements. Güley \cite{guley2014methodological} worked on a city district by recording and analyzing the existing urban data for the analysis of VDP in the use of color on facades. Moussazadeh and Aytug \cite{moussazadeh2018concept} conducted research on the analysis of VDP and Gestalt principles in contemporary museums using a systematic manual detection method that presents an objective attitude to elaborate on the factors of the formal aesthetics. Although those studies present meaningful approaches, they are debatable and impractical for a serial analysis and application process due to their manual, time-consuming and subjective nature \cite{ostwald2009line}. Therefore, various computational methods have emerged to improve facade analysis; they include Hough transformation \cite{ostwald2009line}, numerical fractal analysis \cite{ostwald2010mathematics} and point cloud analysis \cite{balzani2017point}. \

\subsubsection{AI-based studies using facades}

Chalup and Ostwald \cite{ChalupOstwald2010} presented psychological-based measures in visual perception based on various ML models to understand the information processing mechanisms in human beings pertaining to aesthetic perception in the built environment. They predicted that AI could be used for more objective analysis of facades. Keeping this in mind, some of the AI-based studies on facades are segmentation \cite{mathias2016atlas}; window detection \cite{ali2007window,neuhausen2018automatic}; building entrance detection \cite{liu2014entrance}; 3D modeling of facade visuals \cite{simon2011random}; detection of facade, number of floors and windows from aerial images \cite{meixner2010interpreting}; detection of repetitive elements on the front facade \cite{lettry2017repeated}; surface material detection \cite{yang2016towards}; and style classification \cite{mathias2011automatic,yi2020house}. While appreciating these applications, we are aware of the need to go beyond the clearly defined visual problems like window detection, and to identify a way of analyzing complex visual compositions of contemporary facades from an architect’s aesthetic point of view. \

\subsection{Computational analysis of visual aesthetics}

Aesthetics is the study of the judgment of beauty and ways of seeing and perceiving the world \cite{white2011elements}. For the evaluation of aesthetics, visual design elements are analyzed by integrating various computational methods. The measurement of the beauty of images with those methods is called computational aesthetics (CA) \cite{bo2018computational}. Brachmann and Redies \cite{brachmann2017computational} and Tang et al. \cite{tang2019review} give detailed reviews of concepts and methods of experimental CA. Some of the studies are explained in the following subsections, including computational methods with and without ‘learning’, and hint at some of the VDP that are also within the scope of this research. \

\subsubsection{Studies on Computational Aesthetics (CA)}

The effect of the VDP on visual aesthetics is well accepted. In CA, aesthetic evaluation relates with some features, such as “including interesting content, object emphasis, good lighting, color harmony, vivid color, shallow depth of field, rule of thirds, balancing element, motion blur, repetition and symmetry” \cite{yang2019comprehensive} (p.297). Ngo \cite{ngo2001measuring} created a quantitative assessment model for screen formats, which predicts overall measures of 14 VDP. Lin \cite{lin2013development} developed a scale to measure VDP by constructing a structural model using exploratory and confirmatory factor analysis to understand the relationships between VDP; such as balance, contrast, rhythm, and dominance and the perceived ease of use and aesthetics. Some researchers implement Arnheim’s \cite{arnheim1974art} theories on visual weight and balance for the evaluation of the compositional characters of a 2D image. Akkad and Gazimzyanov \cite{al2017automated} presented a method for estimating the balance of a visual scene using the design elements for the calculation of perceptual forces. Li et al. \cite{li2019aesthetic} used visual balance for the quantitative assessment of image aesthetics and calculated the ‘visual center’ by collecting various central points related to the factors, such as color, shape and size of the visual elements. These studies were conducted by analyzing compositional features without ML. \

\subsubsection{AI-based studies in CA}

CA has attracted interest from AI researchers in recent years \cite{tang2019review}. A comprehensive survey on image aesthetic quality assessment, including the review of datasets, various aesthetic tasks and applications using recent AI methods, was carried out by Yang et al. \cite{yang2019comprehensive}. \

Several studies conducted with separate AI motives have already hinted at some VDP. Li and Chen \cite{li2009aesthetic} identified the visual aesthetic quality of paintings by AI in order to construct a relationship between human perception and computer vision, and studied affective factors in human judgments, including ‘color’, ‘composition’, ‘general feeling’ and ‘feeling of the brushstrokes’. Ciesielski et al. \cite{ciesielski2013finding} focused on the discovery of image features associated with the aesthetic value of photographs and abstract images with ML, and their proposed model computed features that captured aspects of certain properties, such as color contrast, harmony, tone, and figure and background. Malu et al. \cite{malu2017learning} constructed a multitasking deep CNNs model, which jointly learns eight aesthetic attributes including balancing elements, color harmony, object emphasis and vivid colors, along with the overall aesthetic score; they then evaluated the scores with the help of the visualization of attribute activation maps. Kim et al. \cite{kim2019neural} worked with the Gestalt principle ‘law of closure’ to see if neural networks have the ability to generalize with regard to this visual phenomenon. Inspired by the principles of art and the harmony of visual elements, Liu et al. \cite{liu2020composition} performed a composition-aware analysis, which looks at mutually dependent visual elements in image aesthetics. \

\begin{figure*}
	\centering
		\includegraphics[width=\textwidth,height=\textheight,keepaspectratio]{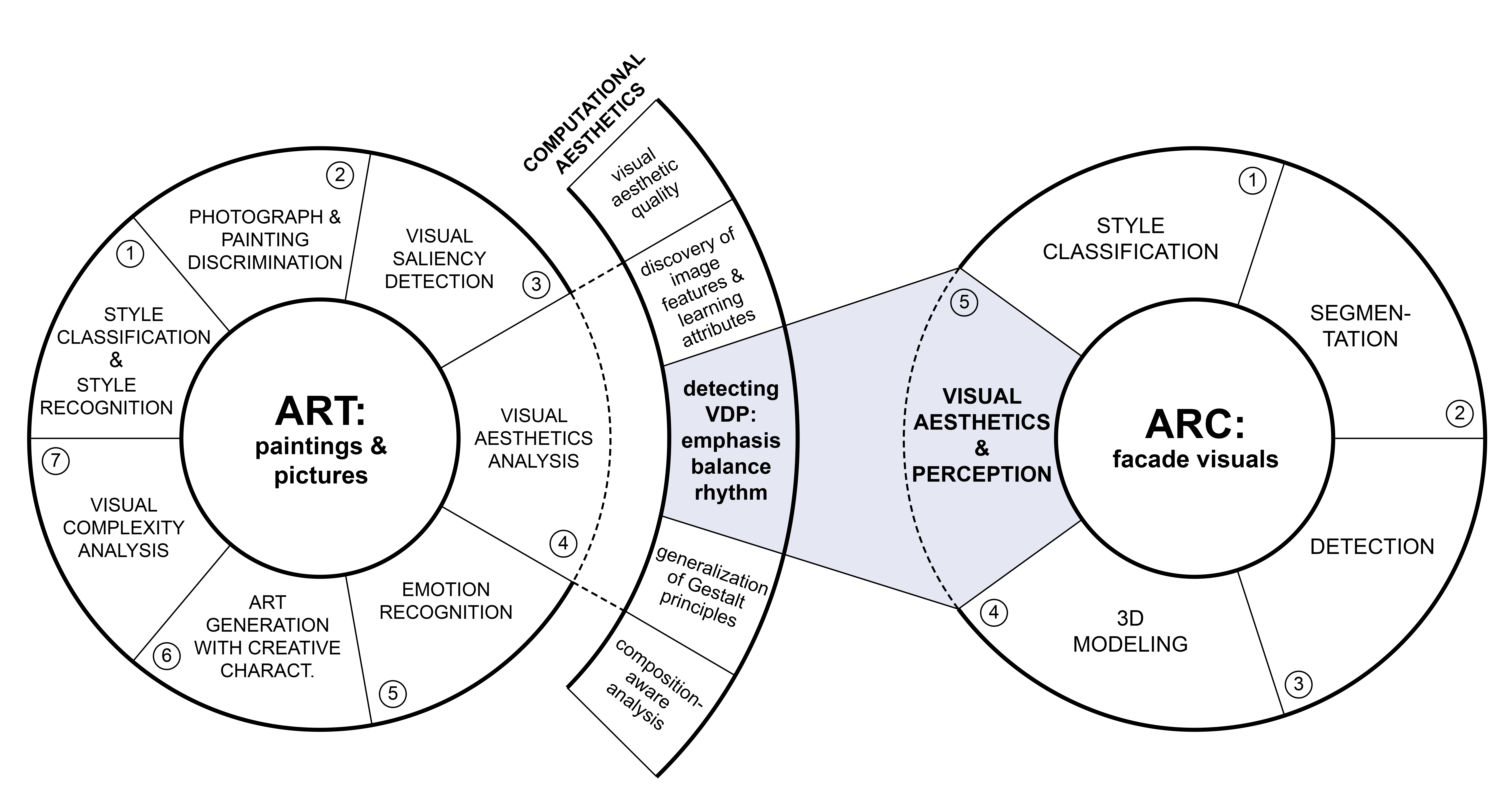}
	\caption{AI-based studies and motivations in art and architecture.}
	\label{F1}
\end{figure*}

To sum up, AI has already been used as a method for visual aesthetics analysis in art but not yet fully in architecture. As an exception, Thömmes and Hübner \cite{thommes2018instagram} look for a correlation between Instagram likes and visual aesthetics with curvature and balance measures for architectural images. Figure \ref{F1} gives an overview of the fundamental AI-based studies and motivations in the ART domain of photographs and paintings, and in the ARC domain of facade images. Our work, matching with the shaded area in the figure, seeks to understand computer performance in the detection of the specific VDP of emphasis, balance and rhythm; not to mention their semantic organization in the design products of art and architecture. To the best of our knowledge, this is the first time that this is being investigated. The proposed AI model here learns the common patterns of these VDP by using a large set of original works; and we hypothesize that this well “trained eye” \cite{nodine1993role} can make judgments on visual compositions that are analogous to the values given by experts. Our research methodology is explained in three main sections: the expression of selected VDP (Section 3); dataset preparation (Section 4); and finally, building and optimizing the Deep Neural Net (CNN) Model (Section 5).  \

\section{Quantifying the VDP}

\begin{figure}
	\centering
		\includegraphics[scale=.75]{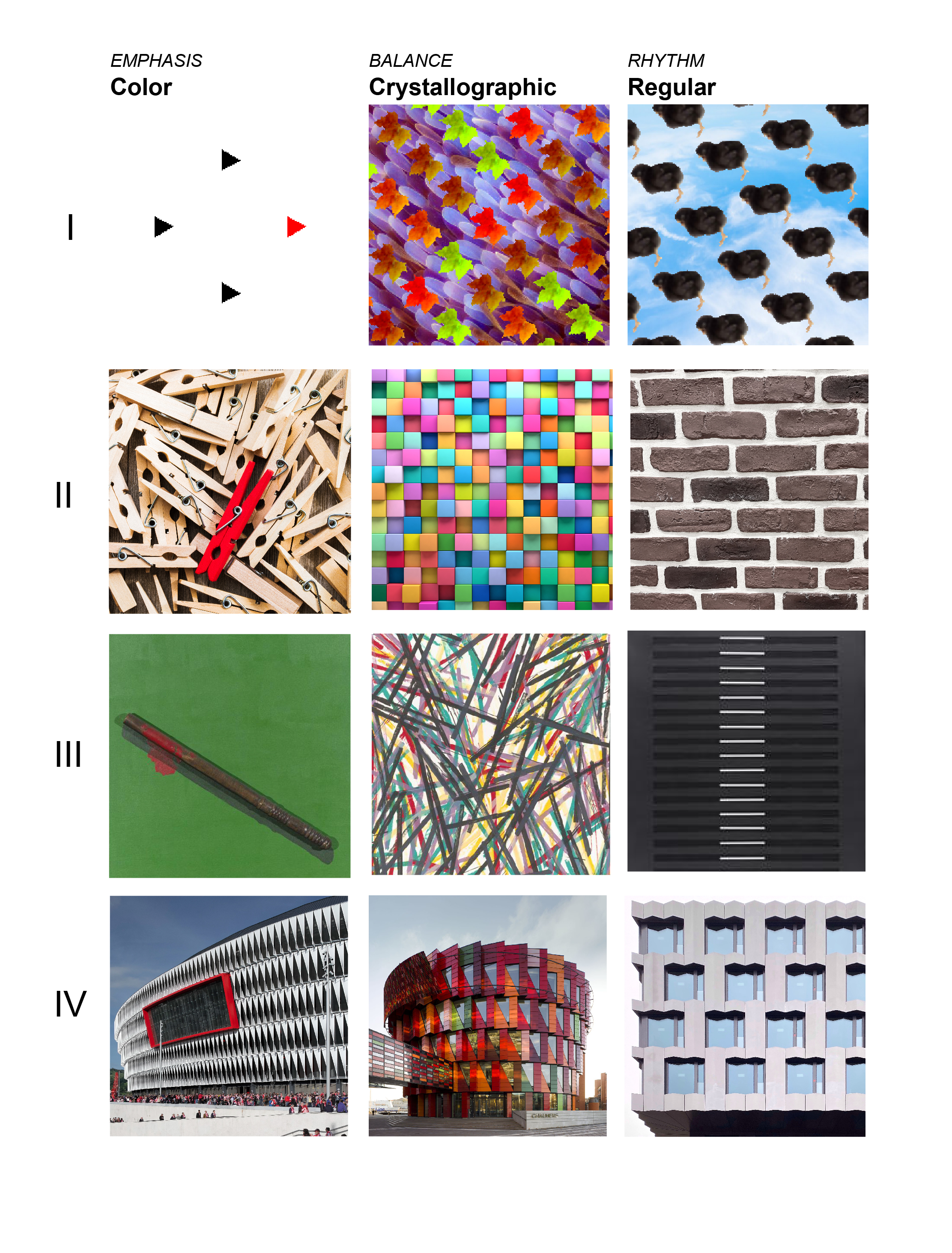}
	\caption{
	Sample data for four different domains. Left-right: \textit{color} (emphasis), \textit{crystallographic} (balance), and \textit{regular} (rhythm). Photography and artwork credits (left to right) are given below.\protect\\\hspace{\textwidth}
    I. Synthetic Image Dataset in Computer Domain.\protect\\\hspace{\textwidth}
    II. Photograph Dataset  in PHT Domain. iStock by Getty Images: /Eblis; /HomePixel; /Miki1988.\protect\\\hspace{\textwidth}
    III. Painting, Print \& Graphic Art Dataset in ART Domain. Collection MCA Chicago: /Roland Ginzel (The Billy Club, 1969); /Chuck Arnoldi (For Decisions and Revisions, 1983); /John Goodyear (Two-sided Movement, 1965).\protect\\\hspace{\textwidth}
    IV. Architectural Facade Dataset in ARC Domain. Arch Daily/Aitor Ortiz (San Mames Stadium, Bilbao); Arch Daily/Tord-Rikard S\"{o}derstr\"{o}m (Kuggen Building, G\"{o}teborg); Instagram/$@$architecturetourist (Technopark, Zurich).\protect\\\hspace{\textwidth}
    }\label{F2}
\end{figure}

The CNN model for detecting emphasis, balance and rhythm is tested on various domains in the visual design field; involving five datasets composed of two sets of synthetic (computer-generated) images, one set of photographs (PHT), one set of paintings, prints and graphic art (ART), and one set of architectural facades (ARC). Sample data from different domains’ datasets are shown in Figure \ref{F2}. \

\subsection{VDP used in the study}

All VDP are independent and have different features and functions \cite{landa2010graphic} in the visual design. For this research, some principles were found to be easier to analytically express and identify as ‘emphasis’; other possibilities, like ‘unity’, were vaguer and rather ambiguous. Three main categories of VDP were selected depending on the art resources \cite{lauer2011design,stewart2006launching}: emphasis, balance and rhythm. The selection was also based on the selected VDP’s sheer number of examples and their shared definitions in the literature. For the supervised deep learning training, three sub-principles (sub-VDP) for each principle; 9 classes in total, were defined: 1 \textit{color}, 2 \textit{isolation} and 3 \textit{shape} for emphasis; 4 \textit{symmetric}, 5 \textit{asymmetric} and 6 \textit{crystallographic} for balance; 7 \textit{regular}, 8 \textit{progressive} and 9 \textit{flowing} for rhythm. \

\textbf{Emphasis} is the creation of dominant elements in a composition. Lauer and Pentak \cite{lauer2011design} define emphasis as a ‘focal point’, which “attracts attention and encourages the viewer to look closer” (p. 56). \textit{Color} emphasis is managed by an element with a contrasting or otherwise distinct color in a composition. \textit{Isolation} is a way of creating emphasis through locating an element apart from other things in the composition. \textit{Shape} emphasis is used when a color-matched element with distinct shape in form or scale appears in a composition. \

\textbf{Balance} is the creation of equality of visual weights in a composition. Arntson \cite{arntson2011graphic} defines balance as “two forces of equal strength that pull in opposite directions, or by multiple forces pulling in different directions whose strengths offset each other” (p. 64). \textit{Symmetric} balance is the reflection of elements within the composition with respect to a centerline or axis. While two sides are not identical, a sense of balance could still be achieved through a clever arrangement of elements in a composition, where an \textit{asymmetric} balance occurs. \textit{Crystallographic} balance is  about repetition and consistency, inspired by the color and shape variations of elements within a composition. An image employing this principle has equal visual weight in all regions. \

\textbf{Rhythm} is the creation of repetition in elements, colors, forms, positive and negative spaces, and textures. Landa \cite{landa2010graphic} defines rhythm as “a sequence of visual elements at prescribed intervals” that develops “a coherent visual flow” from one element to another (p.35). \textit{Regular} rhythm describes a composition that contains the same or similar recurring elements that are usually placed at regular intervals –such as grids. \textit{Progressive} rhythm is the hierarchical change in a group of recurring elements in the composition, like a series of squares getting slightly bigger each time, or a single square transforming into a circle gradually in several frames. \textit{Flowing} is the repetition of wavy lines, bended elements and curved shapes within a composition, which depicts a movement. Best examples are found in organic forms including flowers and clouds. \

\subsection{Setting sub-VDP rules}

In the previous ML models, popular datasets, such as MNIST, ImageNet, PASCAL, and CIFAR-10/100 are used for image classification. They classify the images by the type of “a prominent object or feature,” which indicates “concrete classes” and requires the analysis of the local features. However, it will not be adequate to classify “abstract classes” by “simply considering local features” \cite{stabinger2017evaluation} (p.2767). Therefore, the visual rules and compositional logic of the sub-VDP should be depicted in detail to make a better classifier regarding their semantic contents. \

Working with different types of sub-VDP reveals knowledge of different visual organizations and how the model relates to them. We started building an analytical explanation of all the principles, not to only understand  designing rules in their formal nature but to provide a variety in the dataset –including the generation of synthetic dataset. Thus, we aimed at preventing the model becoming biased towards specific patterns, which would result in a poor generalizer \cite{belem2019impact}. Our approach is similar to Field \cite{field2018illustrated}, who analyzed and visually explained various functions of the design elements and organization of the VDP. \

\begin{figure*}
	\centering
		\includegraphics[width=\textwidth,height=\textheight,keepaspectratio]{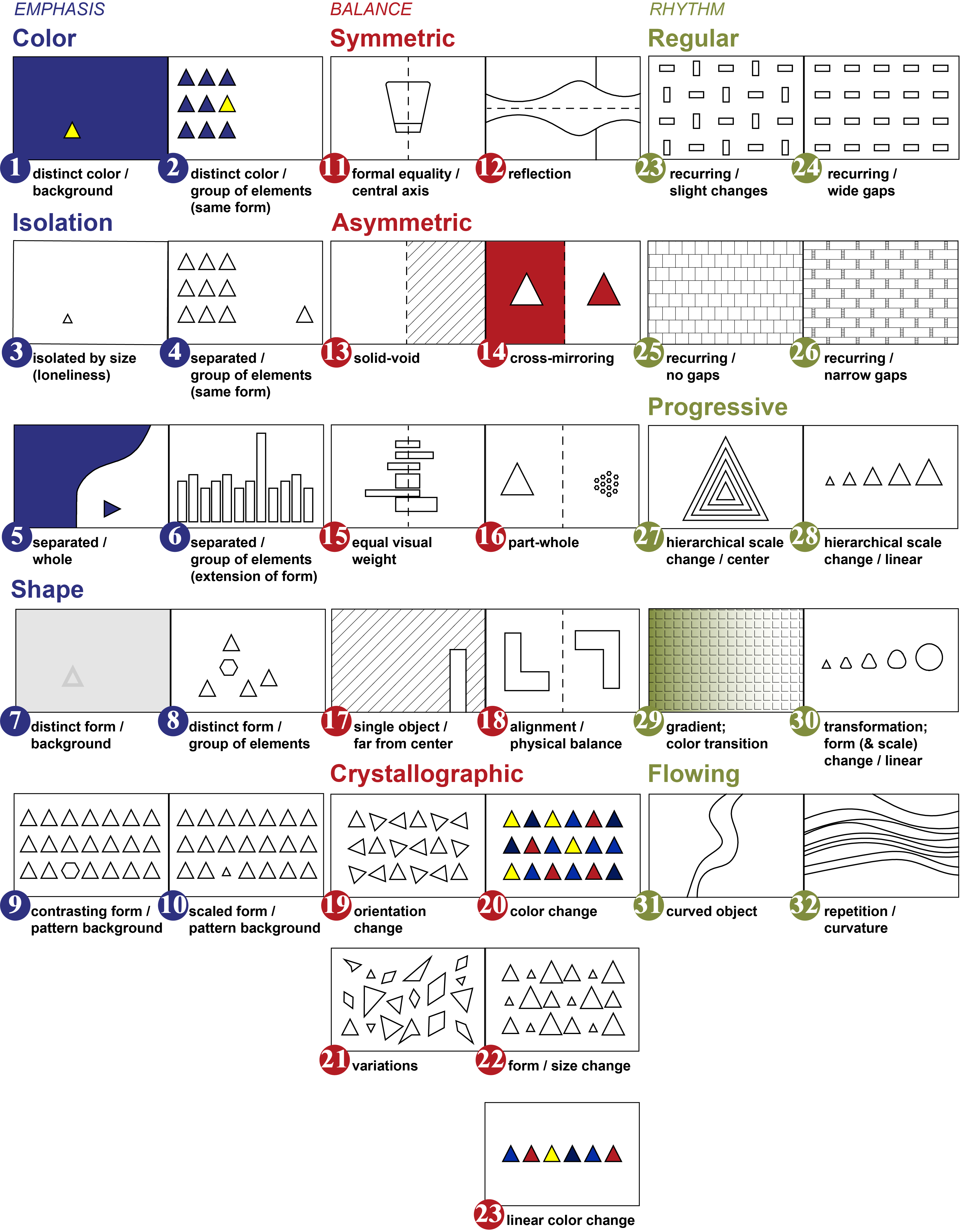}
	\caption{Different types of organization of the visual design elements. Samples of the sub-VDP of emphasis, balance and rhythm: 1–10, 11–22 and 23–32, respectively. For example: for \textit{color}; an element is emphasized with a distinct color on a background (figure–ground relationship) (Rule \#1), or in a group of elements (\#2).}
	\label{F3}
\end{figure*}

We elaborated the selected sub-VDP and extracted 32 design rules in total, based on observations on initially collected data. They are illustrated and explained briefly in Figure \ref{F3}; 2 for \textit{color}; 4 for \textit{isolation}; 4 for \textit{shape}; 2 for \textit{symmetric}; 6 for \textit{asymmetric}; 4 for \textit{crystallographic}; 4 for \textit{regular}; 4 for \textit{progressive}; and 2 for \textit{flowing}. This elementary exercise on quantification would also shed light on what kind of information we expect the computer to glean from the data. \

\section{Curating the dataset}

Most of the studies in the art domain commonly use large-scale and available databases, such as the AVA dataset \cite{murray2012ava}, for visual aesthetic assessment \cite{tang2019review, 10.1007/978-3-319-16178-5_2}. However, for research like ours, which focuses on a design aspect that has not been considered before, a specially curated dataset was necessary – as in the work by Llamas et al. \cite{llamas2017classification}. Since no datasets exist that are labelled according to the compositional rules and VDP, we created a “well-balanced training set” \cite{belem2019impact} (p.288) in the selected domains, including architecture. \

Different approaches exist in dataset preparation for ML studies in the literature. Some studies have harvested stock-image websites like Flickr \cite{hodosh2013framing,zhang2016describing} and several other websites \cite{carballal2018distinguishing,datta2008algorithmic}. Jahanian et al. \cite{jahanian2015learning} collected 120,000 photographs from 500px, an online photo-sharing platform. Wikiart has also been a source \cite{lecoutre2017recognizing}. Web search engines, like Google images \cite{yi2020house} are also used to collect images. \

\begin{figure*}
	\centering
		\includegraphics[width=\textwidth,height=\textheight,keepaspectratio]{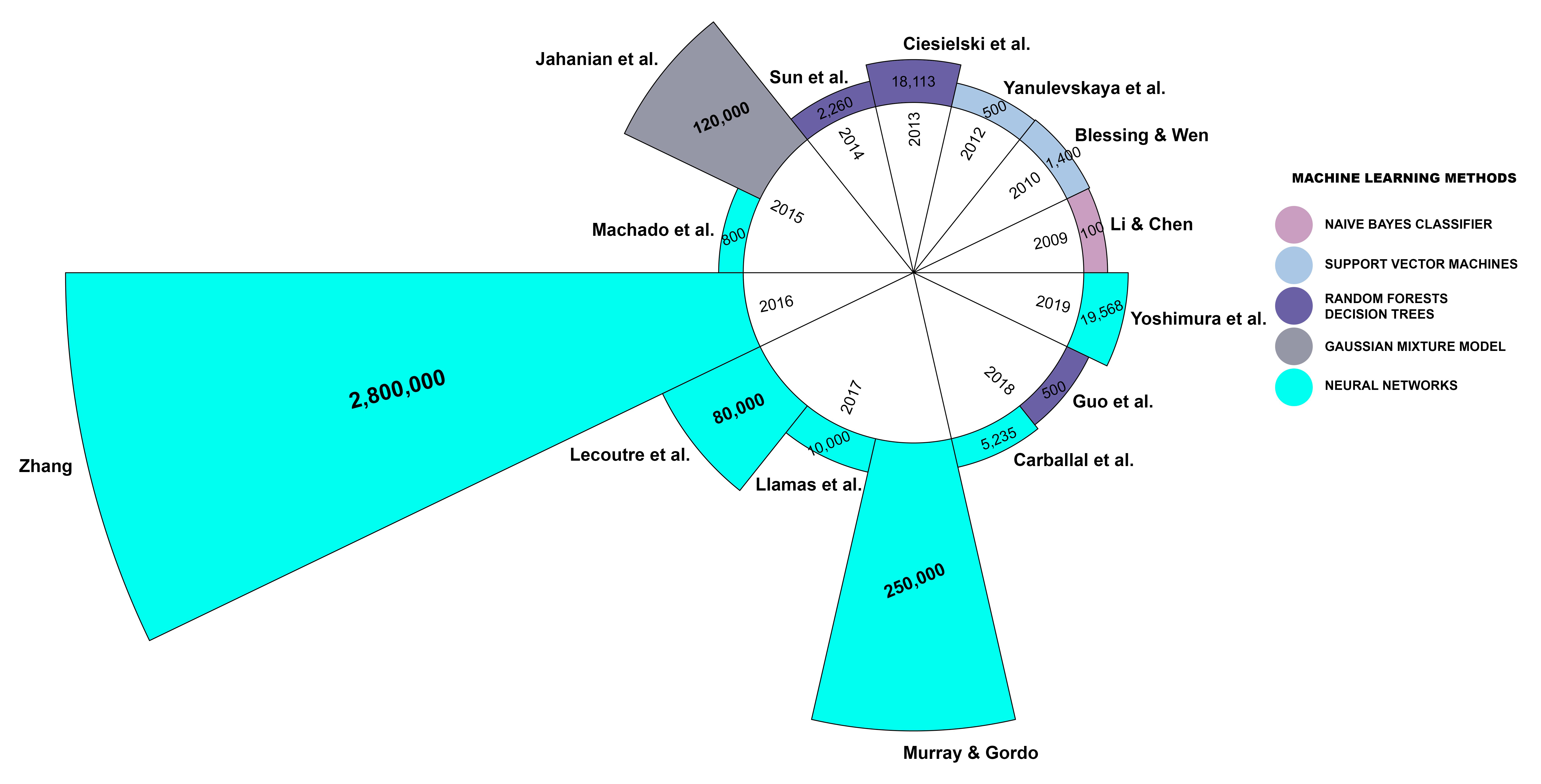}
	\caption{Representing some ML applications and data numbers in the art and architecture field studies, years between 2009-2019.}
	\label{F4}
\end{figure*}

For some ML studies in art and architecture domain, the sizes of the datasets are shown in Figure \ref{F4}. The largest amounts of data are often used in the neural network models. This can be exemplified in art and aesthetics based ML studies as; 2,800,000 \cite{zhang2016describing}, 250,000 \cite{murray2017deep}, and 80,000 \cite{lecoutre2017recognizing}  and in architecture based ML studies as; 19,568 \cite{yoshimura2019deep}, and 10,000 \cite{llamas2017classification}.

To warm up, we first created a computer-generated dataset (in Subsection 4.1). Then, to meet the complexity of the problem, we moved on with the preparation of datasets from multiple domains (Photography in Subsection 4.2; Art in Subsection 4.3; and Architecture in Subsection 4.4). We gathered more than 250,000 visuals by web-crawling through various stock-image and museum websites for selecting proper images. After the final data annotation process given in Subsection 4.5, we ended up with 23,825 labeled images in total for training our CNN model, which is described in Section 5. \

\subsection{Synthetic datasets}

At the beginning of this work, lacking any relevant dataset, we needed a proof of concept to show that the end goal was feasible –before starting the costly process of creating our own dataset. With this motivation we computationally created synthetic datasets, which are called Synthetic Dataset Version 1 (SDV1); consisting of simple geometrical shapes, and Synthetic Dataset Version 2 (SDV2); created as a patchwork of photographs. \

\begin{figure}
	\centering
		\includegraphics[scale=.75]{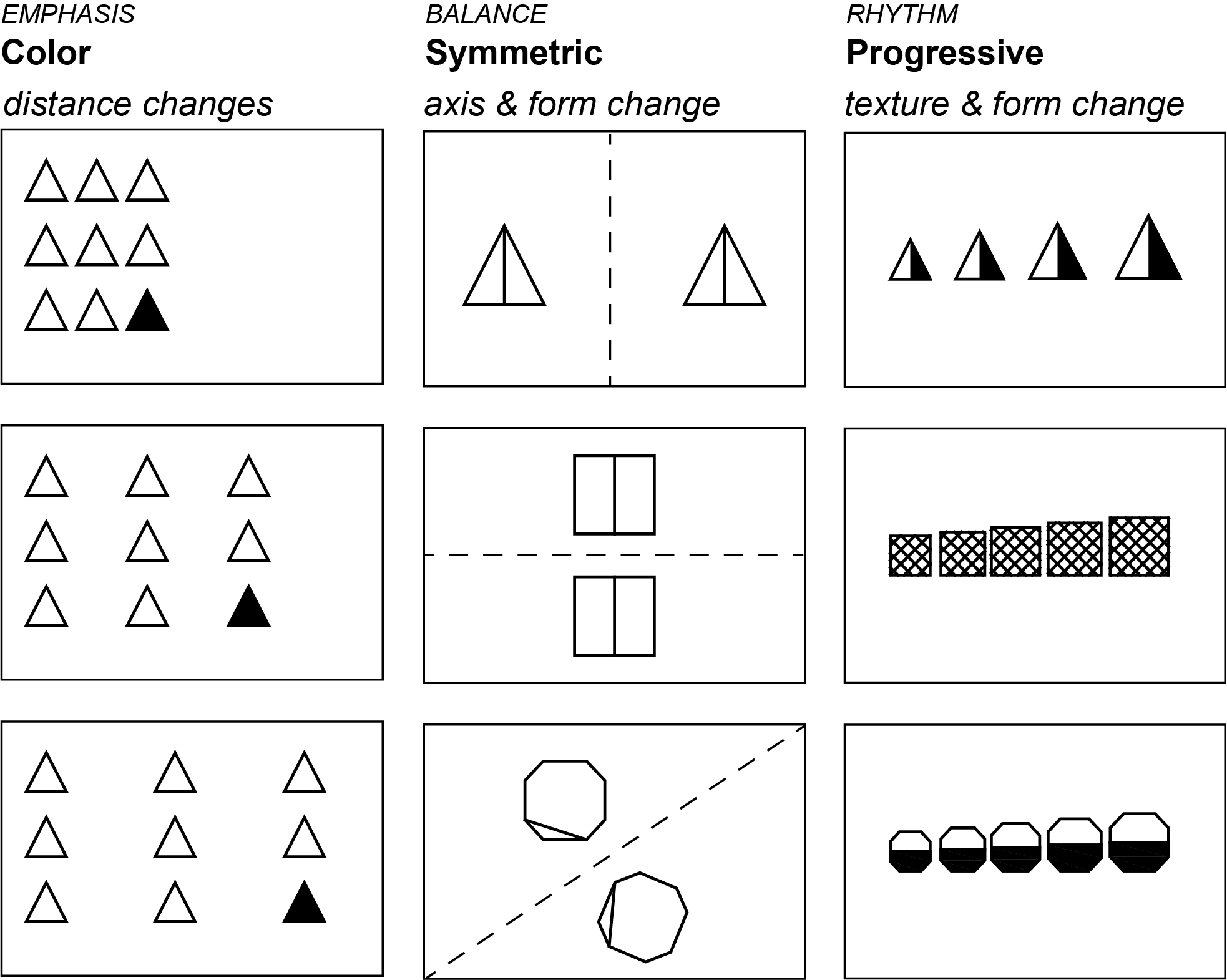}
	\caption{Applied rules in SDV1, showing the changes in parameters such as distance, axis, form, and texture.}
	\label{F5}
\end{figure}

\begin{figure}
	\centering
		\includegraphics[scale=.75]{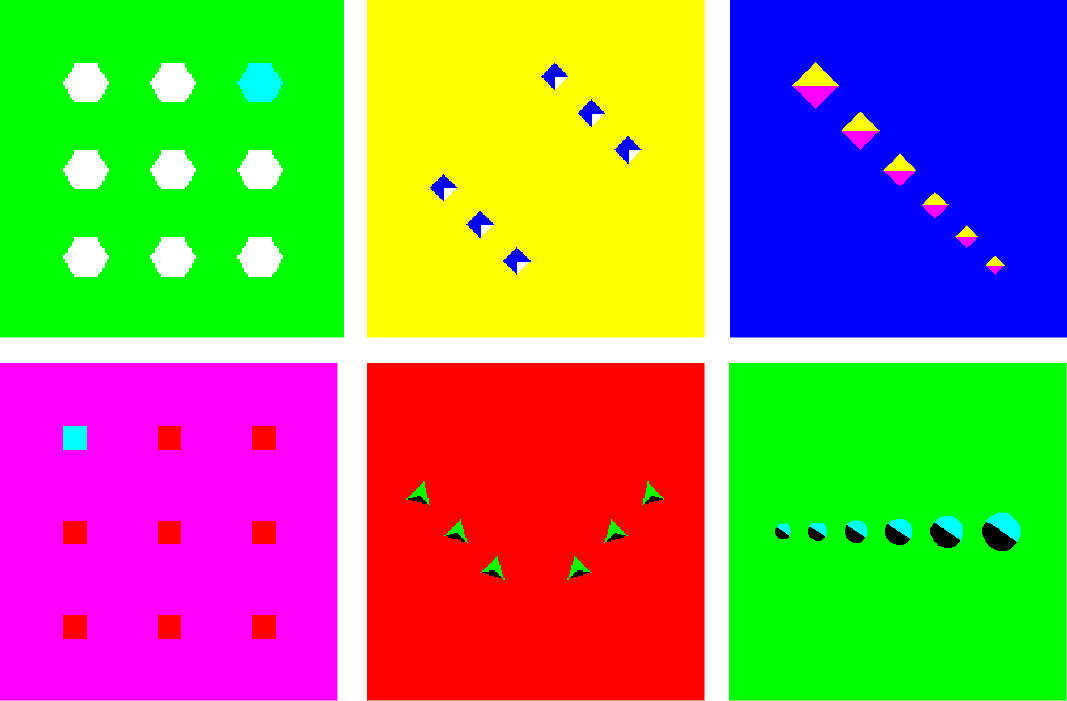}
	\caption{Generated samples from SDV1. Columns from left to right: sample data of \textit{color}, \textit{symmetric} and \textit{progressive}. (Rules \#2, \#12 and \#28 in Figure \ref{F3} are used)}
	\label{F6}
\end{figure}

It is nearly impossible to fully represent a real-life art data due to the enormous amount of variation, which makes developing a hardcoded prediction model very difficult. So, to alleviate the pending shortage in diversity, we referred to the underlying patterns in VDP, and initially parameterized the rules given in Figure \ref{F3} in order to create SDV1. We inspected what features would change, and took a discrete sample of the semi-continuous space of all possible feature combinations. We decided on features such as the orientation of the main axis, distance from the main axis, angle of the shape grid, number of/gaps between columns and rows of the grid and so on, with proper possible values for each of them (Figure \ref{F5}). Even when keeping the set of possible values small in size, thanks to the large number of features we ended up with a space of millions of samples, some depicted in Figure \ref{F6}. This required computing power and time; thus, to make it as fast as possible, we used NumPy – a fast mathematical computation library. To simplify the problem, we implemented functions to create polygons and grids, rotating and translating selected objects, making symmetry according to an axis and so on. Since SDV1 consists of regular polygons, all the calculations are made on the set of essential points and the image is created from these points at the end. \

In SDV1, we generated a subset of 300,000 samples for only three sub-VDP: \textit{color}, \textit{symmetric} and \textit{progressive}, and achieved around 95\% accuracy on both the training and test splits. The results hinted that our dataset was not complex enough and was not a challenge for the network –hence, the model is overfitting. With the insight collected from the SDV1 experiments, and considering the possible weaknesses of the dataset, we moved onto SDV2. \

\begin{figure}
	\centering
		\includegraphics[scale=.75]{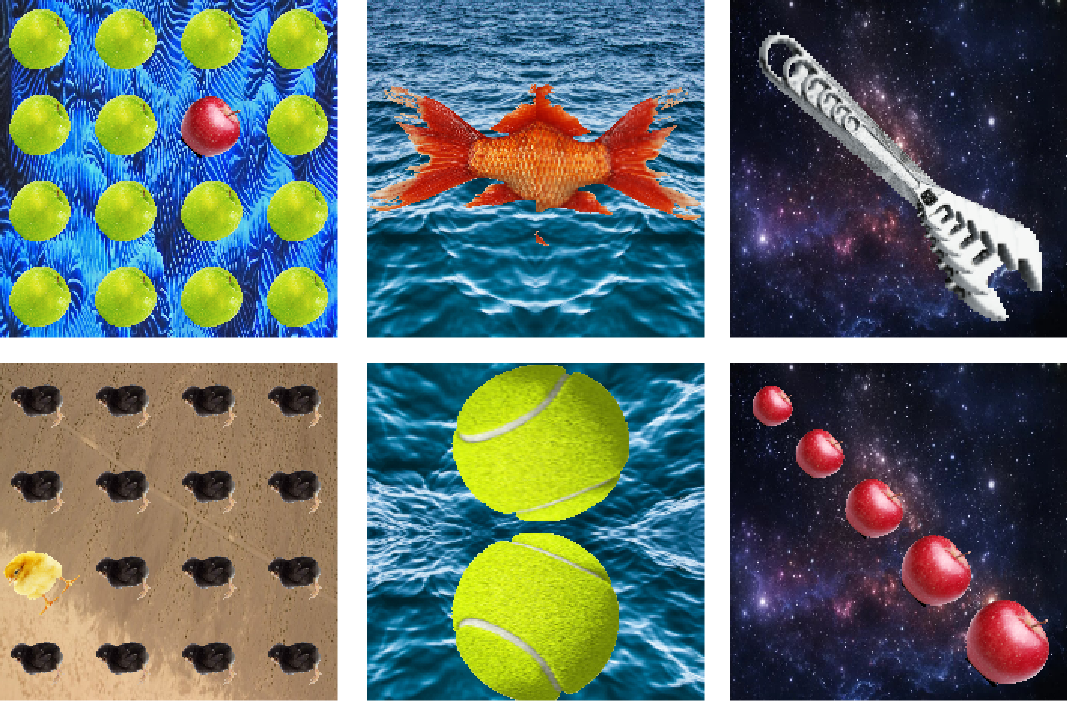}
	\caption{Generated samples from SDV2. Columns from left to right: sample data of \textit{color}, \textit{symmetric} and \textit{progressive}.}
	\label{F7}
\end{figure}

SDV2 was created in an effort to better represent real-life data. To do this, we added a set of selected objects and backgrounds, all of which were photographs. By adding texture we aimed to generalize the dataset, which is harder to memorize compared to the solid shapes in SDV1. SDV2 also contains all the target sub-VDP we selected for the problem. Similar methods were used in the creation of SDV2 resulting in images as shown in Figure \ref{F7}. Training our model with a subset of 900,000 images proved no different than SDV1 in terms of accuracy – around 95\% on both the train and test splits. This indicated that even though SDV2 is more complex in detail, it is still possible to easily train our network on it with very satisfactory results. \

At each step of creating and training on the synthetic datasets, we aimed to manipulate the network to obtain a higher accuracy; however, the CNN model worked effortlessly, and we therefore did not dwell on the network further. Instead, we moved to work on the real datasets. In this preliminary work, by curating SDV1 and SDV2, we saw that it would be possible to teach the defined patterns of the sub-VDP to an advanced network with an appropriate dataset. \

\subsection{Photography dataset}

We first used Google Images, which works with an object-based search algorithm for image retrieval. There, we could not acquire the semantic content, inherent in the source image, from the retrieved image. Another challenge was that, as most of the images are watermarked with the name of the website, they were not eligible for use. There was also an abundance of computer-generated images. Another problem with using a search engine is that it increases the number of data-gathering sources, whereas we prefer to keep the number of different sources as small as possible for unity within the dataset. \

\begin{figure}
	\centering
		\includegraphics[scale=.75]{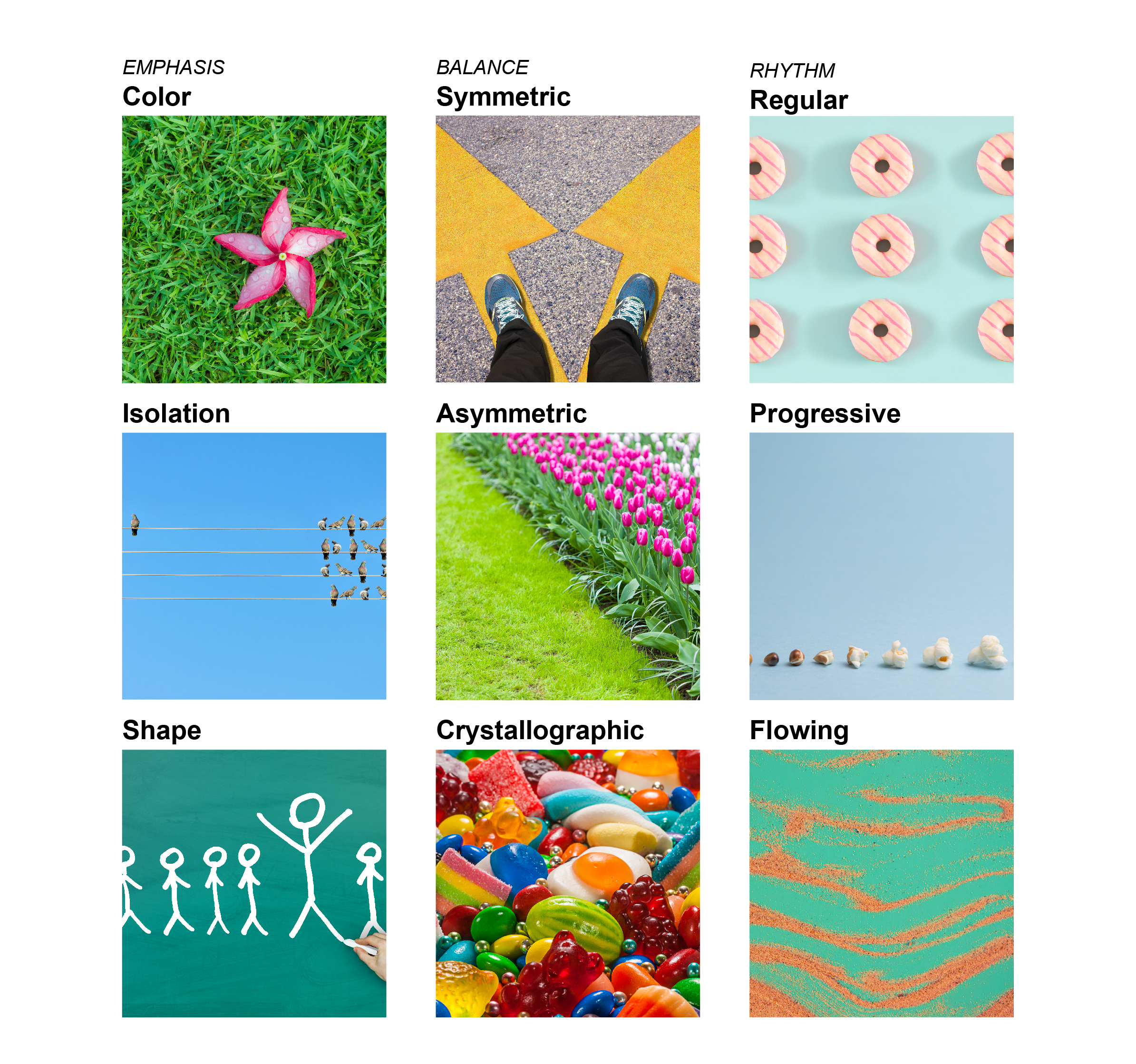}
	\caption{Sample images from the photography dataset. iStock by Getty Images, photography credits (left to right then, top to bottom): /Poravute; /Dontstop; /portishead1; /OgnjenO; /Paket; /Fodor90; /AndreyPopov; /TheCrimsonMonkey; /Liia Galimzianova.}
	\label{F8}
\end{figure}

For this dataset, the majority of the images was collected from stock image websites, such as iStock and the 500px collection of Getty Images. Other websites, such as Shutterstock and Unsplash, were also used. We made a conceptual keyword search in 500px. A search by keyword resulted in additional keywords previously tagged on the image. For example, a label like ‘reflection’ is found to be related to ‘symmetry’, so the images labeled with ‘reflection’ are inspected as well – thereby aiding us in finding new data. On the other hand, this method is limited to the tagged labels of the stock image website. Therefore, both detailed and varying combinations of the existing keywords were utilized in the searches, while related concepts, words and objects were also generated with the analysis of the existing data that could lead to new data. For example, ‘Newton’s cradle’ is an object that does not come up in a search for emphasis; however, it is a perfect example of \textit{isolation} or ‘yin-yang’, which is a concept related to \textit{asymmetric}. Some of the found and generated keywords of the related labels are shown in Table \ref{tbl1}. Sample images from the constructed PHT dataset are shown in Figure \ref{F8}. \

\begin{table*}[t]
\caption{Examples of the keywords, found to be related by the selected sub-VDP. (The keywords collected from Getty Images are given in bold).}\label{tbl1}
\resizebox{0.8\textwidth}{!}{\begin{tabular}{lllllllll}
\multicolumn{9}{l}{\textbf{Keywords of the sub-VDP}}                                                                                                                                                                                  \\
\multicolumn{3}{l}{\textbf{EMPHASIS}}                                                & \multicolumn{3}{l}{\textbf{BALANCE}}                                  & \multicolumn{3}{l}{\textbf{RHYTHM}}                                    \\
\textbf{\textit{color}}                       & \textbf{\textit{isolation}}   & \textbf{\textit{shape}}         & \textbf{\textit{symmetric}}  & \textbf{\textit{asymmetric}} & \textbf{\textit{crystallographic}} & \textbf{\textit{regular}}    & \textbf{\textit{progressive}}    & \textbf{\textit{flowing}}       \\ \hline
\textbf{standing out from the crowd} & \textbf{left behind} & \textbf{individuality} & \textbf{reflection} & \textbf{imbalance}  & \textbf{multi-\textit{color}ed}    & \textbf{repetition} & \textbf{progress}       & \textbf{wavy}          \\
\textbf{alone in a crowd}            & \textbf{missing out} & \textbf{}              & \textbf{}           & \textbf{}           & \textbf{variation}        & \textbf{pattern}    & \textbf{hierarchy}      & \textbf{rippled}       \\
\textbf{rebellion}                   & \textbf{left out}    & \textbf{}              & \textbf{}           & \textbf{}           & \textbf{diversity}        & \textbf{}           & \textbf{reduction}      & \textbf{flow}          \\
\textbf{unique}                      & \textbf{exclusion}   & \textbf{}              & \textbf{}           & \textbf{}           & \textbf{abundance}        & \textbf{}           & \textbf{transformation} & \textbf{swirl pattern} \\
\textbf{}                            & \textbf{loneliness}  & \textbf{}              & \textbf{}           & \textbf{}           & \textbf{\textit{color} change}     & \textbf{}           & \textbf{}               & \textbf{}              \\ \hline
black sheep                          & swarm                & relief                 & butterfly           & puzzle              & floral pattern            & brick               & decay                   & desert                 \\
the ugly duckling                    & still life           &                        &                     & ying-yang           & fabric                    & tile                & metamorphosis           & smoke                  \\
                                     & Newton’s cradle      &                        &                     & jenga               & paints                    &                     &                         & silk                  
\end{tabular}}
\end{table*}

\subsection{Art dataset}

\begin{figure}
	\centering
		\includegraphics[scale=.75]{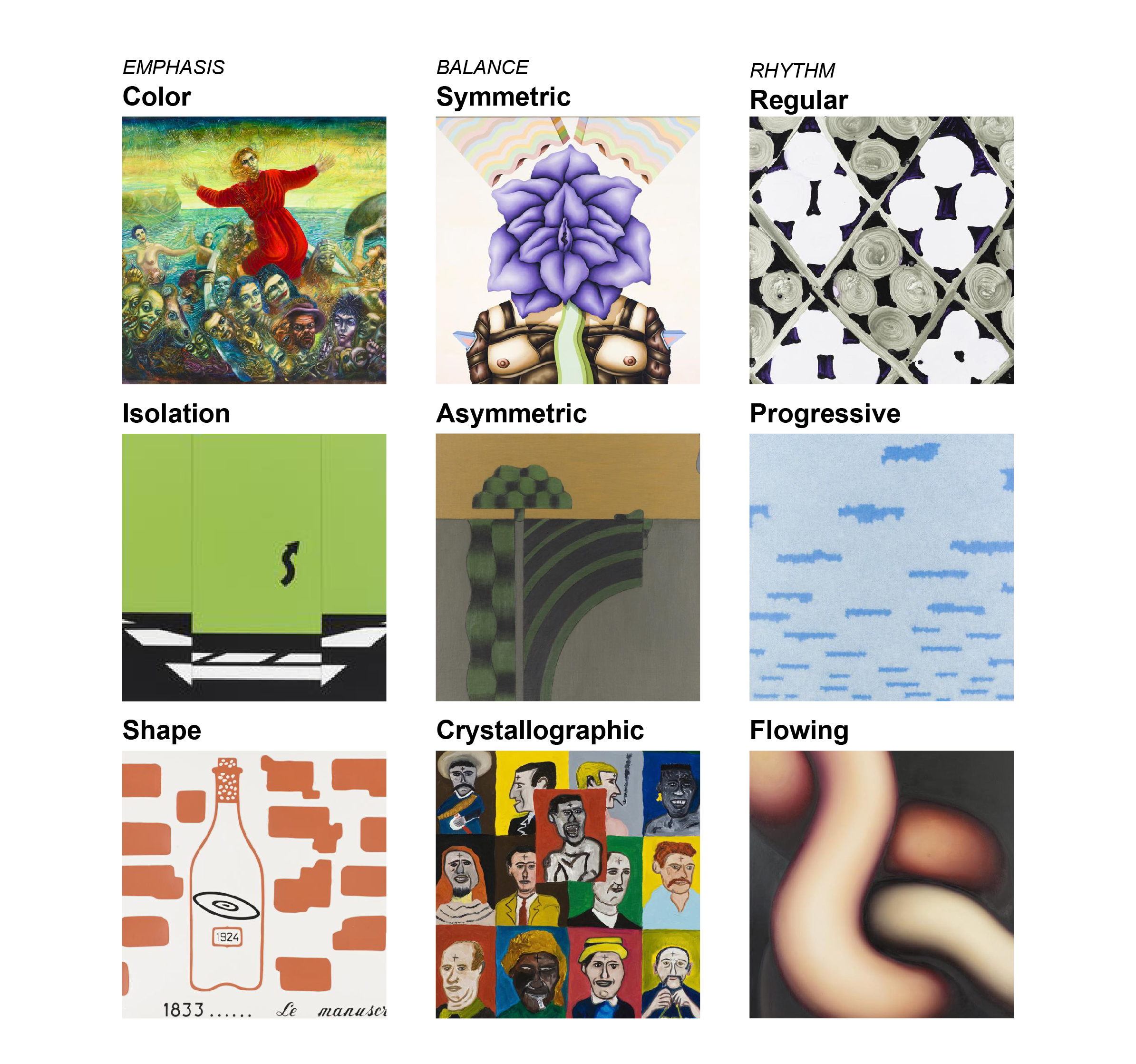}
	\caption{Sample images from the art dataset. Collection MCA Chicago, artwork credits (left to right then, top to bottom): /Alejandro Romero (Procession, 1991); /Robert Lostutter (Untitled (Blue flower), 1972); /Judy Ledgerwood (Thick ‘n’ Thin, 2003); / Allan D'Arcangelo (Landscape I, 1965); /Ben Akkerman (Landscape With Tree (Green-Black-Ochre), 1966–67); /Julia Fish (Cumulous, 1990); /Marcel Broodthaers (1833……Le Manuscrit, 1969–1970); /Rafael Ferrer (Los Diecisiete Aurelianos, 1982); /Frank Piatek (Hierosgamos II (374th tube painting), 1967)}
	\label{F9}
\end{figure}

We focused on contemporary art genres starting from the year 1950, which many art historians consider the end of modern art, or modernism. Genres and styles included optic art, abstract art, pop art and spatialism. We looked for data in 23 contemporary art museum online databases for the selection of (analogue and digital) paintings, prints, works on paper, graphic art, and posters. We avoided collecting photographs and portraits. Based on our observations, artworks have rich visual contents –meaning many of them use more than one sub-VDP with similar visual dominances in their compositions. This led to a more subjective labeling process. Our experience was that \textit{asymmetric} and \textit{crystallographic} samples were common; however, \textit{regular} samples were very few. The digital museum databases and the genres we used are shown in Table \ref{tbl2}. Sample images for the ART dataset are shown in Figure \ref{F9}. \

\begin{table*}[t]
\caption{The total number of data per types collected from various museums.}\label{tbl2}
\resizebox{.6\textwidth}{!}{\begin{tabular}{lllllll}
\textbf{Museum} & \multicolumn{6}{l}{\textbf{Collected data numbers \# per type}}                                                       \\ \hline
\textbf{}       & \textbf{Painting} & \textbf{Print} & \textbf{Work on paper} & \textbf{Graphic art} & \textbf{Poster} & \textbf{TOTAL} \\ \hline
Museum of Fine Arts Boston (MFA)             & 359  & 1932  &     &       &      & 2290  \\
The Museum of Contemporary Art Chicago (MCA) & 303  &       & 668 &       &      & 971   \\
Chrysler Museum of Art                       & 104  &       & 116 &       &      & 220   \\
The Cleveland Museum of Art                  & 278  & 3084  &     &       &      & 3362  \\
Carnegie Museum of Art (CMOA)                & 166  & 730   &     &       &      & 896   \\
Dallas Museum of Art (DMA)                   & 488  & 998   &     &       &      & 1486  \\
Los Angeles County Museum of Art (LACMA)     & 483  & 2337  &     &       &      & 2820  \\
Museum Ludwig                                & 194  & 32    &     &       &      & 226   \\
The Metropolitan Museum of Art (The MET)     & 1541 & 2352  &     &       &      & 3893  \\
Museum of Modern Art (MOMA)                  & 1248 & 1201  &     & 2447  &      & 4896  \\
Museo Nacional Centro de Arte Reina Sofía    & 285  &       & 186 & 104   &      & 575   \\
The Nelson-Atkins Museum of Art              &      & 1015  &     &       &      & 1015  \\
Phoenix Art Museum                           & 197  &       & 186 &       &      & 383   \\
Princeton University Art Museum              & 392  & 322   &     &       &      & 714   \\
Smithsonian American Art Museum (SAAM)       & 1587 & 3200  &     & 3663  &      & 8450  \\
Städel Museum                                & 192  & 10    &     &       &      & 202   \\
Stedelijk Museum Amsterdam                   & 2155 & 4400  &     & 10703 &      & 17298 \\
TATE Modern                                  & 1602 & 11332 &     &       &      & 12934 \\
Victoria and Albert Museum (V\&A)            &      & 333   &     &       &      & 333   \\
Walker Art Center                            & 336  & 1646  &     &       &      & 1982  \\
Whitney Museum of American Art               & 2160 & 6913  &     &       &      & 9073  \\
Worcester Art Museum (VAM)                   & 144  & 2021  &     &       &      & 2165  \\
Yale University Art Gallery                  & 2443 & 7784  &     &       & 1283 & 11510 \\
Moderna Museet                               & 1454 &       &     & 1658  & 133  & 3245  \\
\textbf{TOTAL}  & \textbf{18110}    & \textbf{51682} & \textbf{1156}          & \textbf{18575}       & \textbf{1416}   & \textbf{90939}
\end{tabular}}
\end{table*}

\begin{figure}
	\centering
		\includegraphics[scale=.75]{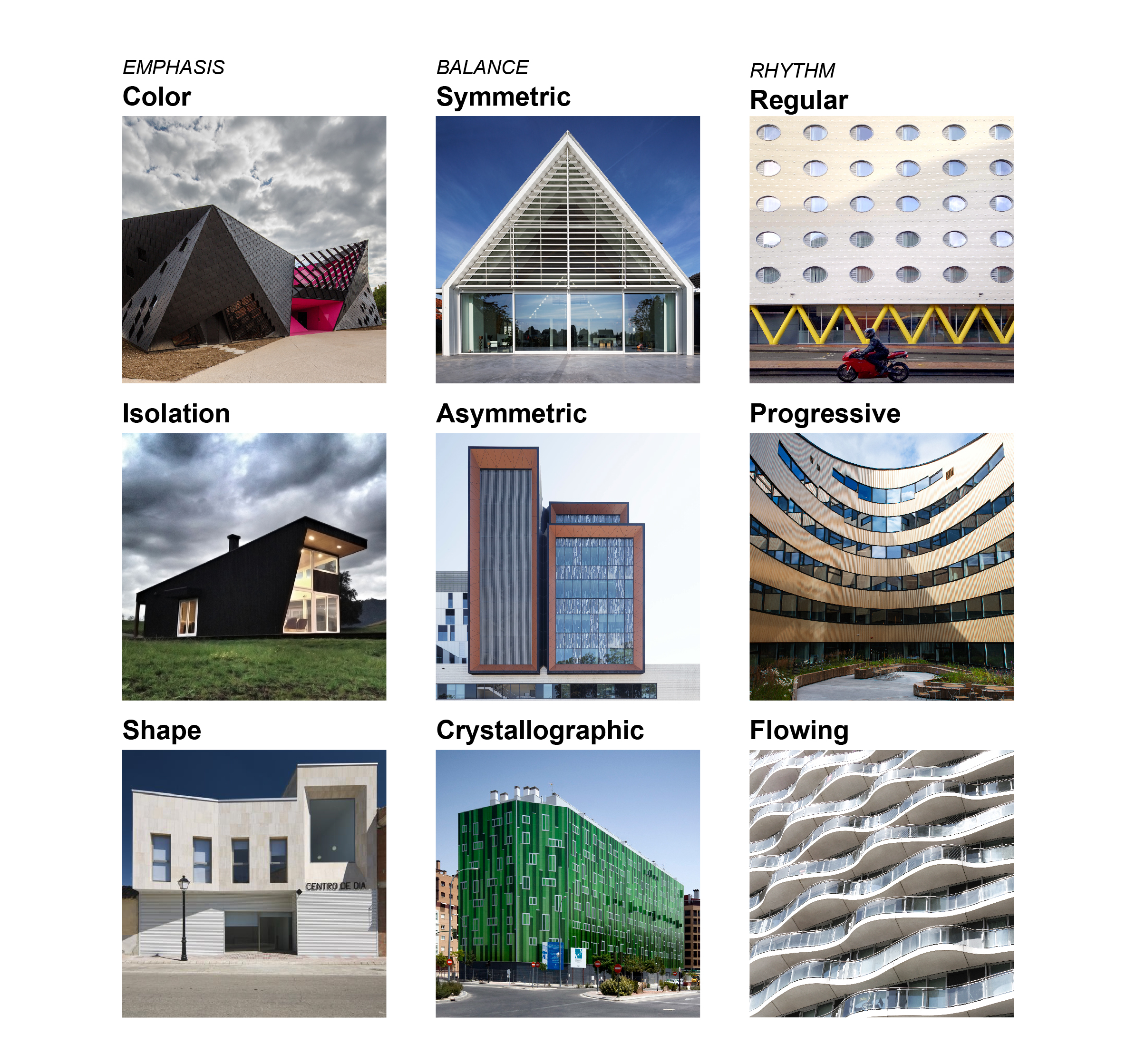}
	\caption{Sample images from the architecture dataset. Photography credits (left to right then, top to bottom): Arch Daily/Florent Michel (Cultural Center, Mulhouse); Arch Daily/René de Wit (Houses, Adaptive Reuse, Rotterdam); Instagram/@architecturetourist (Premier Inn Hotel, Cardiff); Arch Daily/Tomás Alvarez Robledo  (Country House, Chile); Arch Daily/John Gollings  (Northern Beaches Hospital, Australia); Arch Daily/Ieva Saudargaité  (Energy Plant, Trondheim); Arch Daily/Miguel Souto  (Rehabilitation Center, Munera); Arch Daily/SOMOS.Arquitectos (Social Housing, Madrid); Instagram/@architecturetourist (Hoola Building, London).}
	\label{F10}
\end{figure}

\subsection{Architecture dataset}

As Wang \cite{wang2019data} states, our goal was to “datafy” unstructured data; therefore, we confined ourselves to “machine-readable and machine-processable” (p.128) architectural data of facades with evident use of VDP. We used two main web apps for the data collection in the ARC dataset: Instagram and ArchDaily. We only selected contemporary buildings and eliminated historic buildings. Unlike the other datasets, this dataset has variance in styles, functions, architects, and locations. We also selected singular facades instead of building rows. We eliminated building images with statues, cars, urban furniture, humans and other elements that disturbed the visual composition. Also, night and black-and-white photos were not included. We did not add technical drawings, illustrations and renders into the dataset. \

We selected main elevations: front, back, right and left sides, also any perspective scene as a possible view of a building while approaching. Thus, a large number of alternative images could be obtained from a certain building regarding its outer appearance. A front facade with windows on a uniform grid could be labeled as ‘\textit{regular},’ whereas a slight diversion from a direct position would change it as ‘\textit{progressive}’, since the objects appear distorted when farther away. We discarded extreme camera angles like top-down and worm’s eye view. In the work of Thömmes and Hübner \cite{thommes2018instagram}, the classification of architectural images as having a 2D or 3D representation depended on “rotation invariance” (p. 5), meaning if the image is 2D, it lacks a recognizable top or bottom part. In this study, we ignored the evident 3D features, such as depth, which would cause a misinterpretation of VDP. Sample images from the ARC dataset are shown in Figure \ref{F10}. \

\subsection{Annotation of the Dataset}

We collected around 100,000 images for photography, 91,800 images for art and 90,736 images for architecture (282,536 images in total) to inspect and annotate in the final phase. Since we had created the synthetic datasets by defined rules, annotation was not necessary. For other domains, as in the work of Cetinic et al. \cite{cetinic2020learning}, we sought professional support to evaluate the sub-VDP in the images. We consulted a team of five professional architects and five artists, who had equivalent experience in the visual design field. To direct the team members to a more objective evaluation base, the rules and some samples of the nine sub-VDP were presented at the beginning of the process. \

\begin{figure}
	\centering
		\includegraphics[scale=.75]{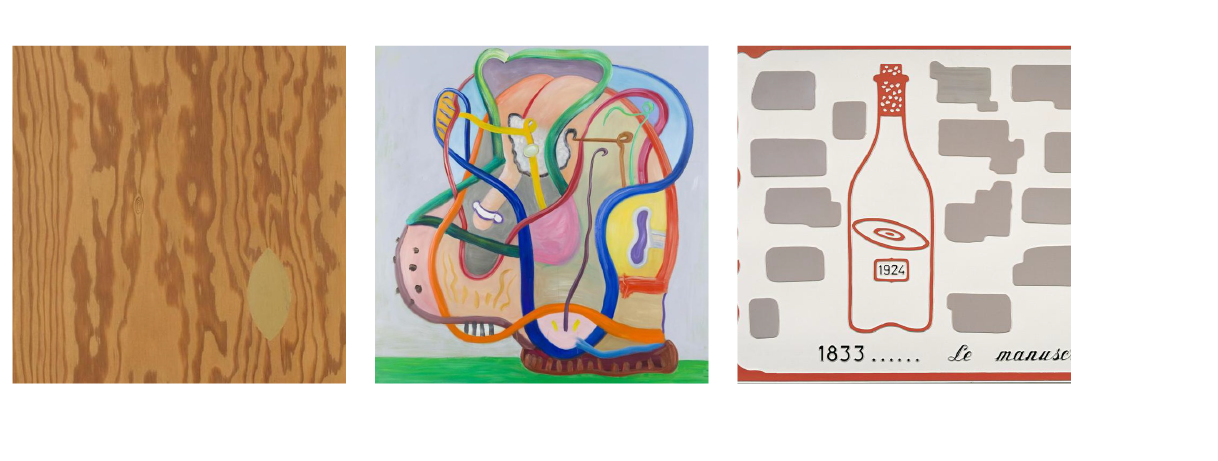}
	\caption{Samples of data; where sub-VDP coexist. Left to right the data have the sub-VDP: \textit{shape}-\textit{flowing}, \textit{crystallographic}-\textit{flowing}, \textit{color}-\textit{shape}. Collection MCA Chicago, artwork credits (left to right): /Sherrie Levine (Untitled (Gold Knots: 1), 1985); /Jim Lutes (Good Teeth and Bad Credit, 1994); /Marcel Broodthaers (1833……Le Manuscrit, 1969–1970).}
	\label{F11}
\end{figure}

We observed that multiple sub-VDP coexist in the data (even in the computer-generated dataset), as in Figure \ref{F11}. Evans and Thomas \cite{evans2012exploring} divide VDP into primary and support principles. They state that “primary principles affect the design as a whole” (e.g. balance), and “support principles affect the internal relationships of a design” (e.g. emphasis and rhythm) (p. 3). Here, we decided to label the image with only the most evident sub-VDP, thereby the data object would belong to one of several pre-defined categories. \

\begin{figure*}
	\centering
		\includegraphics[width=\textwidth,height=\textheight,keepaspectratio]{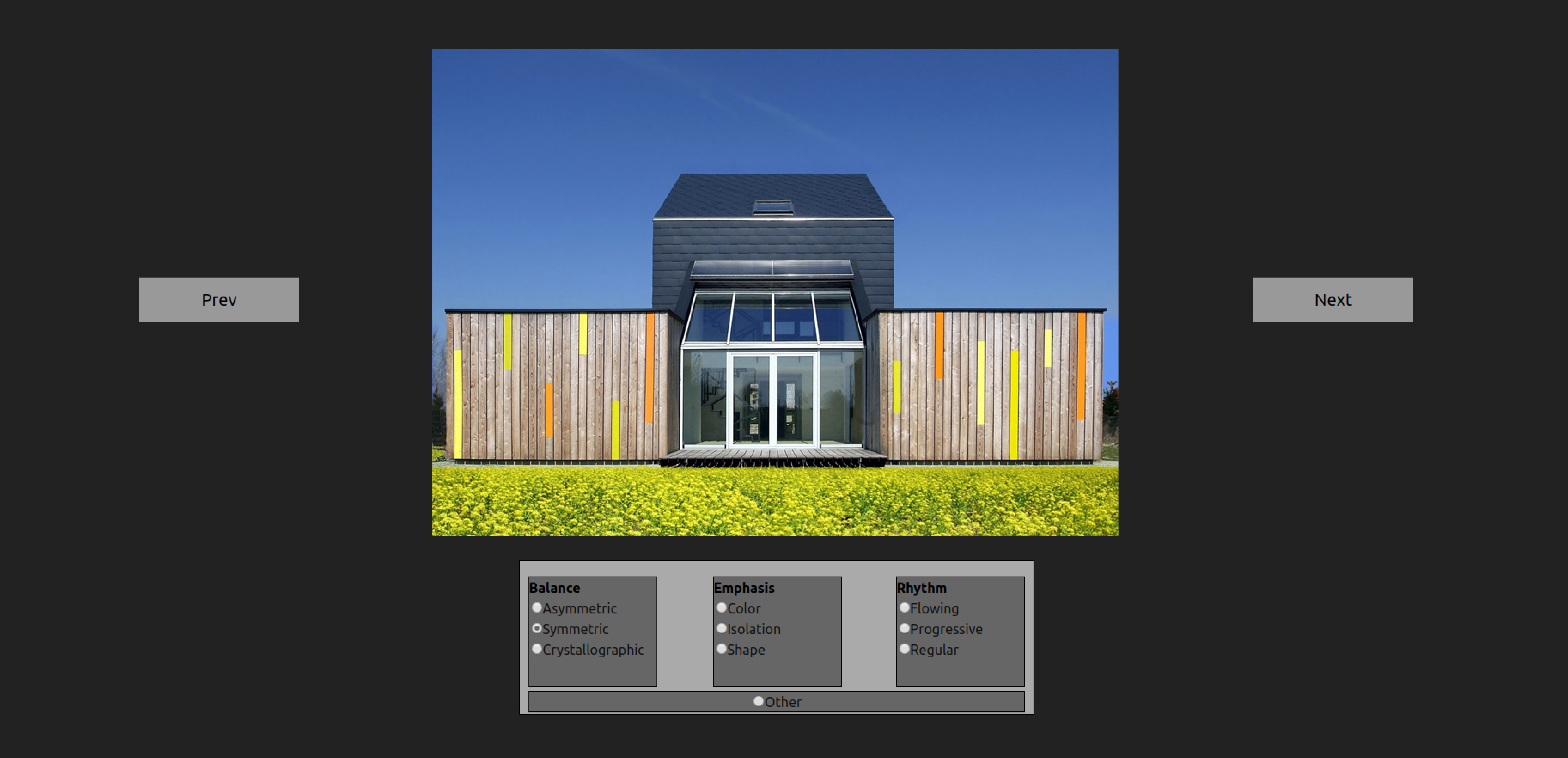}
	\caption{User interface for data annotation with sample data from ARC dataset. Photography credits: Arch Daily/Peter Kuczia (Houses, Pszczyna County).}
	\label{F12}
\end{figure*}

A web app was used for the data annotation process. The platform stores data and has a selection interface (shown in Figure \ref{F12}). The user selects the most apparent sub-VDP, and therefore the label of the data. If the user is unsure or thinks that none of the labels fit the data instance, they click on the ‘other’ option to move to the next one. \

Each team member labeled 25,000 images on average. We made a final elimination after this process with the consensus of the first and second authors. The curated final dataset is comprised of 23,825 labeled images. The data numbers per label are summarised in Table \ref{tbl3}. \

\begin{table*}[t]
\caption{Total data numbers for the curated datasets.}\label{tbl3}
\resizebox{0.8\textwidth}{!}{\begin{tabular}{lllllllllll}
\textbf{Domain}    & \multicolumn{10}{l}{\textbf{Total number of data per each design principle}}            \\ \hline
\textbf{} &
  \multicolumn{3}{l}{\textbf{EMPHASIS}} &
  \multicolumn{3}{l}{\textbf{BALANCE}} &
  \multicolumn{3}{l}{\textbf{RHYTHM}} &
  \textbf{} \\
\textbf{} &
  \textbf{\textit{color}} &
  \textbf{\textit{isolation}} &
  \textbf{\textit{shape}} &
  \textbf{\textit{symmetric}} &
  \textbf{\textit{asymmetric}} &
  \textbf{\textit{crystallographic}} &
  \textbf{\textit{regular}} &
  \textbf{\textit{progressive}} &
  \textbf{\textit{flowing}} &
  \textbf{TOTAL} \\ \hline
Photography (PHT)  & 1,357 & 1,028 & 1,027 & 1,227 & 1,426 & 1,603 & 1,065 & 1,384 & 1,427 & \textbf{11,544} \\ \hline
Art (ART)          & 1,031 & 582   & 701   & 1,033 & 1,062 & 1,040 & 693   & 997   & 997   & \textbf{8,136}  \\ \hline
Architecture (ARC) & 271   & 269   & 550   & 390   & 550   & 550   & 550   & 550   & 465   & \textbf{4,145} 
\end{tabular}}
\end{table*}

\section{Building the CNN model for detecting the sub-VDP}

We initiated our experiments on the simpler synthetically created datasets, which are explained in Subsection 4.1. We started with common CNN models, such as VGG19 \cite{simonyan2014very} and ResNet50 \cite{he2016deep} to gauge the feasibility of and to obtain a proof of concept for our hypothesis that the computer would be able to quantify VDP from a carefully curated dataset of images in the second stage. \

\begin{figure*}
	\centering
		\includegraphics[width=\textwidth,height=\textheight,keepaspectratio]{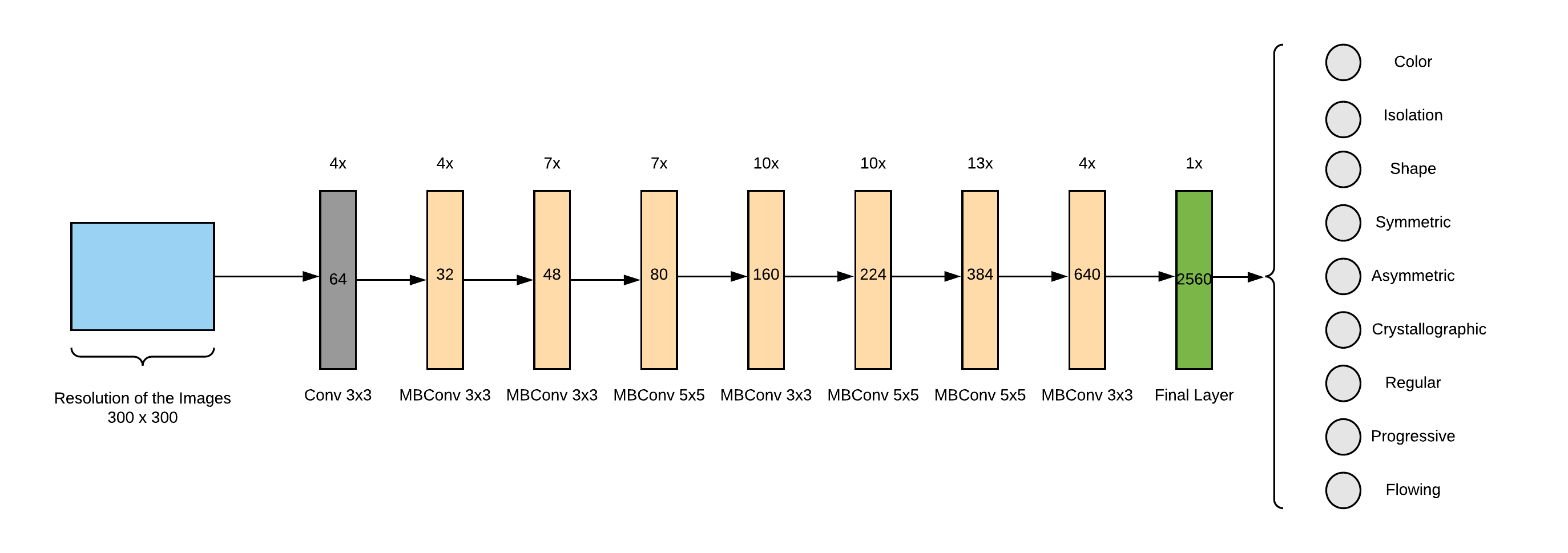}
	\caption{The diagram of our prediction network. The numbers in the boxes are the output channel sizes of the blocks. Yellow boxes are mobile convolution blocks introduced in \cite{DBLP:journals/corr/abs-1801-04381}. The grey box is a standard convolution, the green one is a 1x1 convolution followed by average pooling and a fully connected layer. The number of times the block is repeated are given above the boxes. Below the boxes are the name of the blocks and kernel sizes.}
	\label{F13}
\end{figure*}

Our network has the structure, shown in Figure \ref{F13}. The parameters of a CNN model are the weights of the convolutional and fully connected neural units in the network. These parameters are optimized during training. In addition to parameters of the model, hyperparameters define the architecture in a CNN network. Depth (i.e. the number of layers in the network), width (i.e. the number of filters or number of output channels or neurons in each layer), and resolution of input images are among the important hyperparameters that need to be tuned. These hyperparameters are given in the figure. \

In our initial experiments, we utilized earlier well-known architectures in the deep learning field: VGG19 and ResNet50. Having been fine-tuned over the synthetic dataset, the network model easily fit the data with 95\% accuracy on both train and test sets, because the dataset lacked the rich diversity of a real dataset. Hence, the CNN model could not learn much from the synthetic dataset; however, this did help us to see that it is possible to teach the computer to recognize the nine sub-VDP. We then moved to a more modern classification network: EfficientNet-B7 \cite{tan2019efficientnet}, which is a high-capacity deep learning model. It consumes less time and memory complexity compared to other modern architectures, yet achieves very similar accuracy values. Architectural details of EfficientNet-B7 are given in Table \ref{tbl8}. Conv is standard convolutional layer and MBConv is a slightly more sophisticated MobileBlock \cite{DBLP:journals/corr/abs-1801-04381}. Resolution is the size of an image at each stage accordingly, channels are the number of channels of the output at each stage, repeat is how many of the corresponding operator is used at that stage, AvgPooling is average pooling and FC is fully connected layer for classification. Further details can be found in the original paper \cite{tan2019efficientnet}.\

\begin{table}[t]
\caption{Architectural details of EfficientNet-B7.}\label{tbl8}
\centering
\resizebox{.45\textwidth}{!}{\begin{tabular}{cccc}
\textbf{Operator}        & \textbf{Resolution} & \textbf{Channels} & \textbf{Repeat} \\ \hline
Conv3x3                  & 300 x 300           & 64                & 4               \\
MBConv, k3x3            & 150 x 150           & 32                & 4               \\
MBConv, k3x3            & 150 x 150           & 48                & 7               \\
MBConv, k5x5            & 75 x 75             & 80                & 7               \\
MBConv, k3x3            & 37 x 37             & 160               & 10              \\
MBConv, k5x5            & 18 x 18             & 224               & 10              \\
MBConv, k5x5            & 18 x 18             & 384               & 13              \\
MBConv, k3x3            & 9 x 9               & 640               & 4               \\
Conv 1x1, AvgPooling, FC & 9 x 9               & 2560              & 1              
\end{tabular}}
\end{table}

EfficientNet is not a single network, but a family of networks ranging from EfficientNet-B0 to EfficientNet-B7, and it is possible to increase the number of networks at the end. Indeed, the whole idea of EfficientNet is to scale up CNNs in a smart way so that each resource provided to the network increases the accuracy as much as possible. The base model, EfficientNet-B0, is a very similar model to MobileNetV2 \cite{DBLP:journals/corr/abs-1801-04381}, albeit slightly bigger and with some extra features. \linebreak EfficientNet-B1, which is based on B0, uses twice as many resources as B0 does; and the scaling of depth, width and resolution from B0 to B1 is done in such a way that, with those constraints, the accuracy is maximized. In our work we used the biggest vanilla network version, which is the EfficientNet-B7 CNN model. It was pretrained on the large ImageNet \cite{deng2009imagenet} dataset. For each experiment, we used transfer learning to fine-tune the model towards features of the different data domains.

We use PyTorch \cite{NEURIPS2019_9015} as our deep learning framework and all of our experiments run on this framework. During the optimization for the supervised classification, we used the EfficientNet-B7 that was pre-trained on ImageNet, but without freezing any layer parameters. In other words, we use the pre-trained model as a starting point and then fully optimized the network. All images we used in optimization have a resolution of 300 x 300. For the optimization we used stochastic gradient descent, starting with a learning rate of 0.0256 and exponentiating it every 2.4 epochs. The loss function used in our optimization process was the multi-class cross entropy loss. During optimization of the model, the training dataset was split into train and validation set by 90\%-10\% and the best hyperparameters were found using this validation set. Photography, art and architecture dataset are trained for 37, 89 and 40 epochs respectively. We share the overall numerical results in detail in Section~\ref{sec:Results}. \

\subsection{Dedicated data augmentation in optimization}

Data augmentation is based on the idea of applying different transformations to the training image data in order to both increase the number of data instances and to improve the variability in the data so that the network model is exposed to and learns from different scenarios within the data distribution. Hence, data augmentation is a crucial process in the proper training of a deep neural network and is widely used by the deep learning community. However, as the datasets we created in this work are different from object classification or detection datasets, one has to carefully pick the augmentations to be applied. Augmentations like perspective or warping deformations may damage the information in some classes like \textit{symmetric} or \textit{asymmetric}, and are therefore best avoided. Similarly, we did not use jittering, since jittering the colors may damage \textit{color} emphasis. For augmentation, we flipped the images in x and y axis and rotated the images 90 or 270 degrees. In addition, we applied two types of brightness augmentation that are not commonly used but were developed specifically for this work. After converting the images into LAB color space, we applied the procedures of ‘tweaking global brightness’, which corresponds to increasing or decreasing the brightness of the image by a small amount, and ‘adding a brightness gradient’, which corresponds to tweaking the luminance of the image on a randomly selected axis from dark to light. 

\begin{figure*}
    \centering
    \begin{tabular}{cccc}

      \includegraphics[width=35mm]{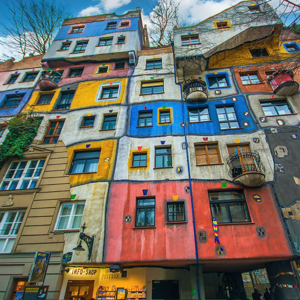} &   \includegraphics[width=35mm]{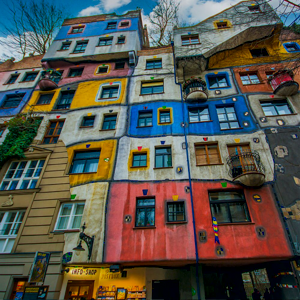} &
      \includegraphics[width=35mm]{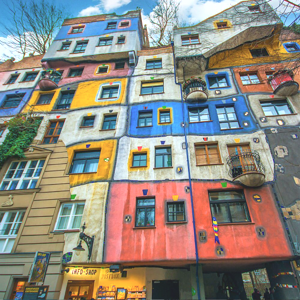} &
      \includegraphics[width=35mm]{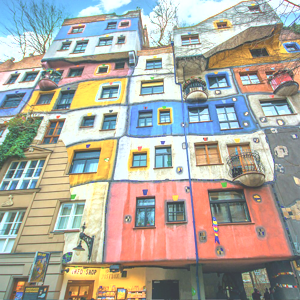} \\
        Original & GBT Sample 1 & GBT Sample 2 & GBT Sample 3 \\[6pt]
        
      \includegraphics[width=35mm]{augs/test.png} &   \includegraphics[width=35mm]{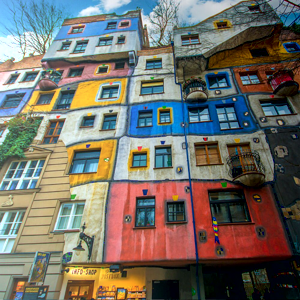} &
      \includegraphics[width=35mm]{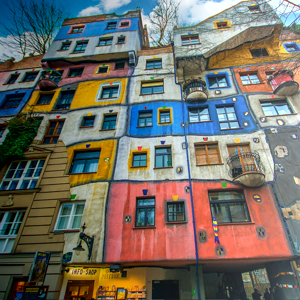} &
      \includegraphics[width=35mm]{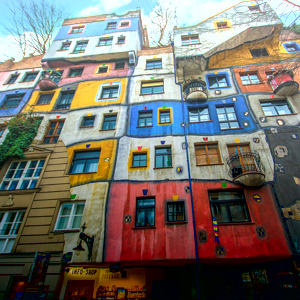} \\
        Original & BG Sample 1 & BG Sample 2 & BG Sample 3 \\[6pt]

    \end{tabular}
    \caption{The figure provides examples of our brightness augmentations. GBT stands for “Global Brightness Tweak” and BG stands for “Brightness Gradient”. Both of our augmentations work in LAB color space, as seen from the figure GBT tweaks the overall brightness up or down and BG creates a gradient in a direction from dark to light. Photography credit: Instagram/$@$kyrenian (Hundertwasserhaus, Vienna, Austria).}
    \label{F_AUG}
\end{figure*}

We provide examples of the described augmentations in Figure \ref{F_AUG} to reflect the gradual effects of both ‘tweaking global brightness’ and ‘brightness gradient’. All augmentations are executed with random sampling. In this way, the CNN model is randomly exposed to a rich set of lighting variations in terms of brightness scale changes towards obtaining both “brighter” and “darker” scenes, as well as varying the overall illuminance grading of the picture in different directions. Similarly, the flipping and rotation augmentations introduce robustness towards certain  geometric transforms in the visual composition. These operations improve on the generalization capability of our CNN model. After the data augmentations, and before training the network, we normalize the images with mean and standard deviation values of the ImageNet dataset. \

\section{Results and Discussion}
\label{sec:Results}
The main purpose of this research was to see if our dedicated CNN model could recognize and classify the underlying VDP in visual compositions, as humans can over various different domains. We modified the EfficientNet model and trained five different models within three different datasets: PHT, ART and ARC. \

\subsection{Quantitative Results}

In the first four cases we trained the model on nine labels of the sub-VDP, but for the last one we targeted the main categories (emphasis, balance, and rhythm) and used only three labels. In two models (Models 1 and 3) we trained the model within different domains; in the remaining three models (Models 2, 4 and 5) we discarded the domains and tested our models with the data from different domains combined. All models and data numbers used for all domains in the train and test sets are summarized in Table \ref{tbl4}. \
 
\begin{table*}[t]
\caption{Total data numbers for all domains in five cases.}\label{tbl4}
\resizebox{0.8\textwidth}{!}{\begin{tabular}{lllllllllll}
\textbf{Model} & \textbf{Domain}   & \multicolumn{9}{l}{\textbf{Total number of data in ‘train-test’ set per each design principle}}           \\ \hline
               &                   & \multicolumn{3}{l}{EMPHASIS}      & \multicolumn{3}{l}{BALANCE}       & \multicolumn{3}{l}{RHYTHM}        \\
 &
   &
  \textbf{\textit{color}} &
  \textbf{\textit{isolation}} &
  \textbf{\textit{shape}} &
  \textbf{\textit{symmetric}} &
  \textbf{\textit{asymmetric}} &
  \textbf{\textit{crystallographic}} &
  \textbf{\textit{regular}} &
  \textbf{\textit{progressive}} &
  \textbf{\textit{flowing}} \\ \hline
1              & PHT               & 1,307-50  & 978-50    & 977-50    & 1,177-50  & 1,376-50  & 1,553-50  & 1,015-50  & 1,334-50  & 1,377-50  \\
               & ART               & 981-50    & 532-50    & 651-50    & 983-50    & 1,012-50  & 990-50    & 643-50    & 947-50    & 947-50    \\
               & ARC               & \textbf{221-50}    & \textbf{219-50}    & \textbf{500-50}    & 340-50    & 500-50    & 500-50    & 500-50    & 500-50    & 415-50    \\ \hline
2              & PHT \& ART \& ARC & 2,509-150 & 1,729-150 & 2,128-150 & 2,500-150 & 2,888-150 & 3,043-150 & 2,158-150 & 2,781-150 & 2,739-150 \\ \hline
3              & PHT               & 210-50    & 210-50    & 210-50    & 210-50    & 210-50    & 210-50    & 210-50    & 210-50    & 210-50    \\ 
               & ART               & 210-50    & 210-50    & 210-50    & 210-50    & 210-50    & 210-50    & 210-50    & 210-50    & 210-50    \\
               & ARC               & 210-50    & 210-50    & 210-50    & 210-50    & 210-50    & 210-50    & 210-50    & 210-50    & 210-50    \\ \hline
4              & PHT \& ART \& ARC & 630-150   & 630-150   & 630-150   & 630-150   & 630-150   & 630-150   & 630-150   & 630-150   & 630-150   \\ \hline
5 &
  PHT \& ART \& ARC &
  \multicolumn{3}{l}{2,820-450 (for emphasis)} &
  \multicolumn{3}{l}{2,820-450 (for balance)} &
  \multicolumn{3}{l}{2,820-450 (for rhythm)} \\ \hline
\end{tabular}}
\end{table*}

The table shows that in Model 1, the train set is composed of the total data (in various numbers: see Table \ref{tbl3}) separated by domain and label. In Model 2, the training dataset is instead composed of the total data (in various numbers) separated by label but not domain. We checked the lowest numbers of data existing in the domains in order to get an equalized dataset among the principles and took the \textit{isolation} data in ARC (219), rounded to 210, as a base. In Model 3, 210 data instances per domain and label were used. In Model 4, 630 (210 * 3) instances per label were used without any domain separation. For the last model, no. 5, we examined the total data numbers of domains for each main principle; the lowest count was ‘emphasis’ in the ARC domain: 940 data instances (221 + 219 + 500, as seen in bold in Table \ref{tbl4}). With taking the equal amount from other domains; 2,820 (940 * 3) instances per main design principle label were used, without any domain separation. For the test set, 50, 150 and 450 instances were used for Models 1 and 3, Models 2 and 4, and Model 5 respectively (Table \ref{tbl4}). \

Visual design composition most likely involves more than one design principle with varying functions \cite{evans2012exploring}. Therefore, instead of judging only the top accuracy score of the computer model, we consider the top three accuracy scores, so that it is possible to evaluate secondary and tertiary predictions of the data. As expected, the model’s top three scoring classes are very likely to include the correct labels, when compared to considering only the top score label. The top accuracy ranged from 56\% to 77\% for all experiments, whereas the top three accuracies ranged from 80\% to 93\%. The accuracies of all predictions are given in Table \ref{tbl5}. \

\begin{table}[t]
\caption{Top three accuracies of five fully-trained EfficientNet models.}\label{tbl5}
\resizebox{.45\textwidth}{!}{\begin{tabular}{llllll}
\textbf{Model} & \textbf{Train domain} & \textbf{Test domain} & \textbf{Top1} & \textbf{Top2} & \textbf{Top3} \\ \hline
1 & PHT               & PHT & \textbf{0.74} & 0.87 & 0.93 \\
  & ART               & ART & 0.71 & 0.81 & 0.87 \\
  & ARC               & ARC & 0.60 & 0.76 & 0.85 \\ \hline
2 & PHT \& ART \& ARC & PHT & \textbf{0.75} & 0.87 & 0.92 \\
  &                   & ART & 0.67 & 0.81 & 0.89 \\
  &                   & ARC & 0.58 & 0.74 & 0.86 \\ \hline
3 & PHT               & PHT & \textbf{0.65} & 0.80 & 0.86 \\
  & ART               & ART & 0.58 & 0.74 & 0.84 \\
  & ARC               & ARC & 0.56 & 0.72 & 0.81 \\ \hline
4 & PHT \& ART \& ARC & PHT & \textbf{0.65} & 0.78 & 0.85 \\
  &                   & ART & 0.59 & 0.76 & 0.85 \\
  &                   & ARC & 0.56 & 0.71 & 0.80 \\ \hline
5 & PHT \& ART \& ARC & PHT & \textbf{0.76} & 0.93 & 1.0   \\
  &                   & ART & 0.75 & 0.94 & 1.0   \\
  &                   & ARC & 0.75 & 0.93 & 1.0  
\end{tabular}}
\end{table}

For a random prediction of nine labels, the expected accuracy without any learning would be $11.\overline{1}\%$; however, the results indicate that the model shows a much better learning performance considering the challenging nature of the experiment. As expected, the accuracy rates are higher for PHT domain, followed by ART and ARC domains. For us, this domain specific condition links with the number and quality of data in the domains. We trained the CNN model with 11,094 samples for PHT, 7686 for ART, and 3695 for ARC (when separated by domain in Model 1). On the other hand, most of the images in the PHT domain have a clear representation of the sub-VDP. Especially marketing photos, selected for this domain (Subsection 4.2), convey the visual messages directly when a VDP related theme is targeted at. However, in the art and architecture data, we find much more complicated visuals, due to the sophistication of the artwork or integration of the structural elements on a facade, such as joints, frames, and installments. \

\begin{figure}
    \centering
    \begin{tabular}{c}
        \includegraphics[scale=.33]{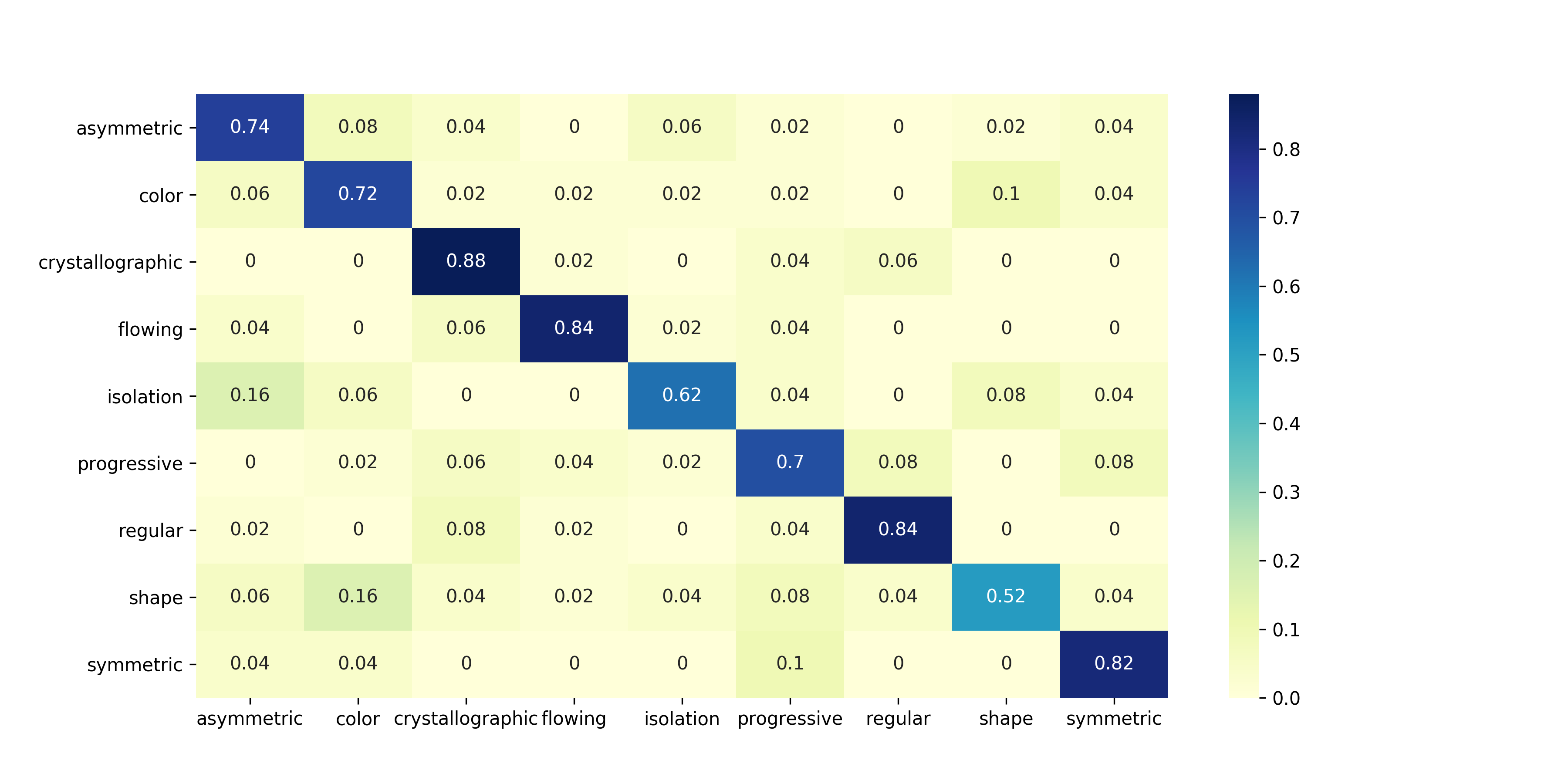} \\
        Photography (PHT) \\[6pt]
        \includegraphics[scale=.33]{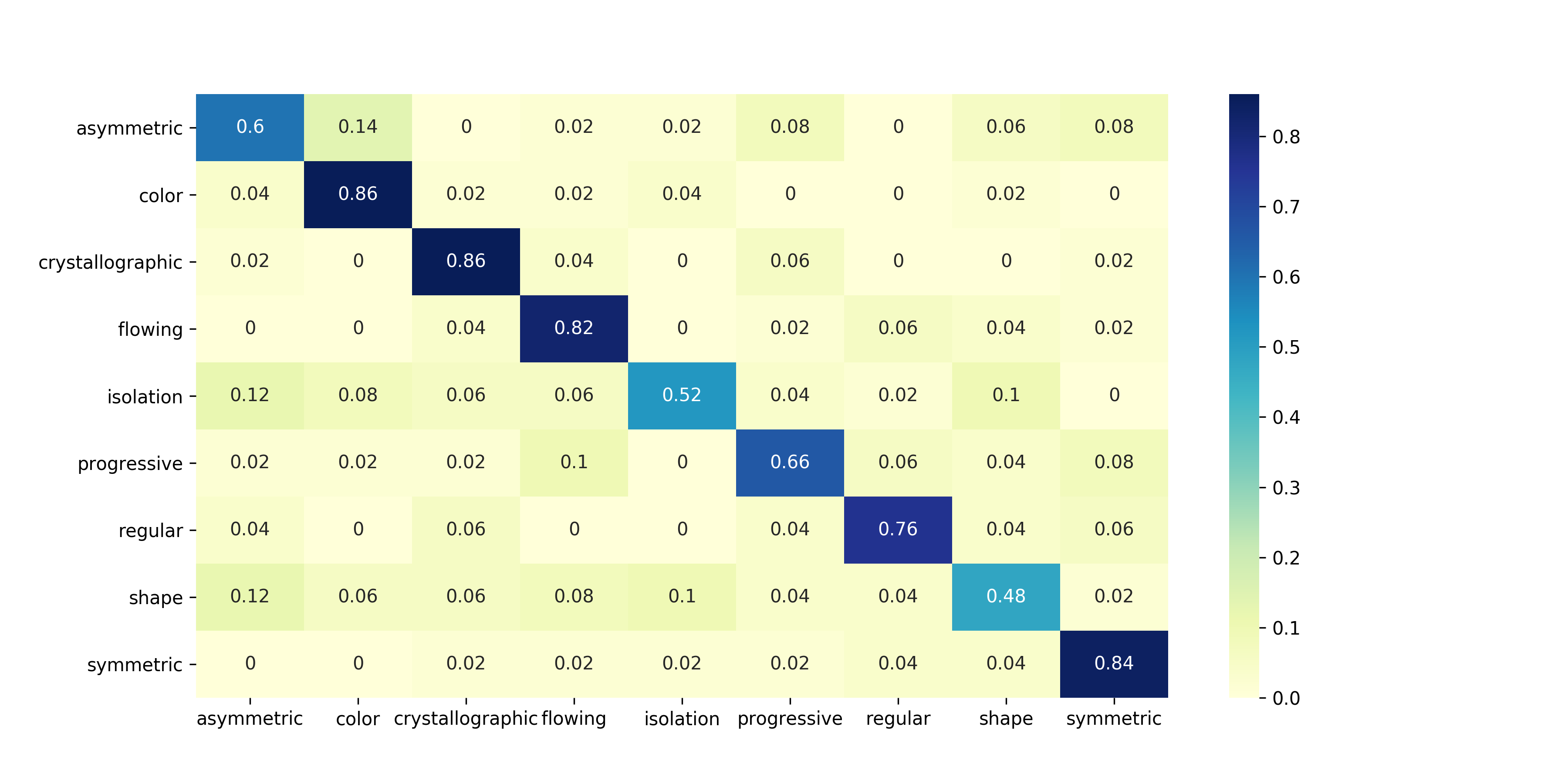} \\
        Art (ART)\\[6pt]
        \includegraphics[scale=.33]{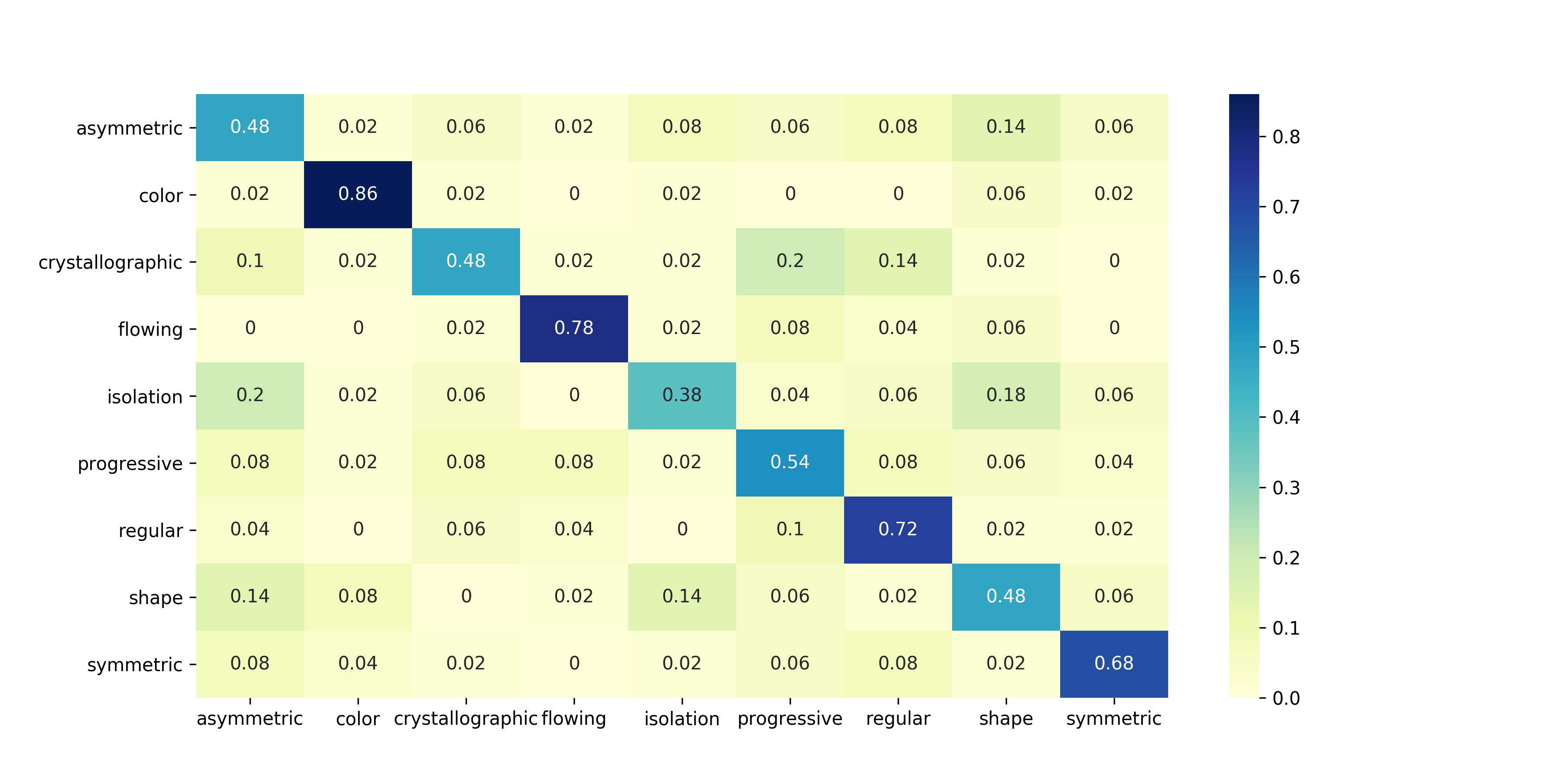} \\
        Architecture (ARC)
        \end{tabular}
    \caption{Normalised confusion matrices for PHT, ART and ARC domains in Model 1. Vertical: predicted labels. Horizontal: actual labels.}
    \label{F14}
\end{figure}

Confusion matrices are used to evaluate the model performance by observing how frequently the image of a label is classified as belonging to another \cite{lindsay2020convolutional}. Normalised confusion matrices for all domains of Model 1 are given in Figure \ref{F14}. The three best performances per labels are \textit{flowing} (rhythm), \textit{color} (emphasis) and \textit{symmetric} (balance); and three worst performances per labels are \textit{shape} (emphasis), \textit{asymmetric} (balance) and \textit{isolation} (emphasis). It is noteworthy that the accuracy rates of a sub-VDP change among the domains. \textit{Color}, i.e., has a better performance in the ART and ARC domains, which is related with the differing visual expressions of sub-VDPs in the domains. \

When the correctly predicted labels are analyzed in the confusion matrices, it is seen that the initial predictions during dataset preparation are coherent with the results. For example, \textit{color} has better rates than the others, as we observed that the color data is easier to classify for us humans as well. Our model obtained higher rates in detecting \textit{symmetric}, which verified the hypothesis of Stabinger and Rodriguez-Sanchez \cite{stabinger2017evaluation} that symmetry detection is “on the border of what current CNN architectures can solve” (p. 2771). \textit{Crystallographic}, \textit{flowing} and \textit{regular} also achieved higher accuracy rates, showing that the model learns to capture their relevant formal characteristics from the data. \

Worst performances (or best as well) cannot be directly or merely related with the numerical non-dominance of the principle in the data; \textit{asymmetric}, i.e., is represented by lots of images, but not satisfactorily detected. We could link this mostly to the annotation procedure of the images. We know that \textit{color} is explicitly emphasized in PHT domain, by the Rules \#1 and \#2, but it mostly coexists with \textit{shape} in the images (with Rule \#9 or \#10). A sample, being identified with more than one rule at the same time, has led to failure in prediction (for top 1) in a specific domain. Since we only give one label to each data, we sometimes had to ignore another sub-VDP. The increase in the accuracies from top 1 to top 3 predictions demonstrate that our model quantifies and supports this knowledge in that these principles mostly co-occur. Another problem arises when two or more sub-VDP have similar compositional structures. \textit{Isolation} and \textit{asymmetric} is an example for that (Rules \#5 and \#14), so they are generally confused with each other in all domains. The highest rates of falsely predicted labels based on the confusion matrices are shown in Table \ref{tbl6}. \

\begin{table}[t]
\caption{The highest rates of falsely predicted labels, based on the confusion matrices of Models 1, 2, 3 and 4.}\label{tbl6}
\resizebox{.45\textwidth}{!}{\begin{tabular}{lllll}
\textbf{Model} & \textbf{Domain} & \multicolumn{3}{l}{\textbf{Highest number of falsely predicted labels}} \\ \hline
\textbf{}      & \textbf{}       & \textbf{Actual}   & \textbf{Predicted (False)}   & \textbf{Number \#}   \\ \hline
1 & PHT & \textit{shape}            & \textit{color}       & 8  \\
  &     & \textit{isolation}        & \textit{asymmetric}  & 8  \\
  & ART & \textit{asymmetric}       & \textit{color}       & 7  \\
  &     & \textit{isolation}        & \textit{symmetric}   & 6  \\
  &     & \textit{shape}            & \textit{asymmetric}  & 6  \\
  & ARC & \textit{crystallographic} & \textit{progressive} & 10 \\
  &     & \textit{isolation}        & \textit{asymmetric}  & 10 \\
  &     & \textit{isolation}        & \textit{shape}       & 9  \\ \hline
2 & PHT & \textit{shape}            & \textit{color}       & 7  \\
  &     & \textit{asymmetric}       & \textit{shape}       & 6  \\
  & ART & \textit{asymmetric}       & \textit{shape}       & 7  \\
  &     & \textit{asymmetric}       & \textit{color}       & 6  \\
  &     & \textit{isolation}        & \textit{asymmetric}  & 6  \\
  &     & \textit{isolation}        & \textit{color}       & 6  \\
  & ARC & \textit{isolation}        & \textit{asymmetric}  & 10 \\
  &     & \textit{isolation}        & \textit{shape}       & 8  \\
  &     & \textit{crystallographic} & \textit{regular}     & 8  \\
  &     & \textit{shape}            & \textit{isolation}   & 8  \\ \hline
3 & PHT & \textit{isolation}        & \textit{asymmetric}  & 9  \\
  &     & \textit{crystallographic} & \textit{regular}     & 7  \\
  &     & \textit{progressive}      & \textit{symmetric}   & 7  \\
  & ART & \textit{symmetric}        & \textit{asymmetric}  & 8  \\
  &     & \textit{shape}            & \textit{isolation}   & 7  \\
  & ARC & \textit{crystallographic} & \textit{regular}     & 11 \\
  &     & \textit{isolation}        & \textit{asymmetric}  & 10 \\
  &     & \textit{isolation}        & \textit{shape}       & 10 \\ \hline
4 & PHT & \textit{asymmetric}       & \textit{color}       & 7  \\
  &     & \textit{asymmetric}       & \textit{isolation}   & 6  \\
  &     & \textit{color}            & \textit{isolation}   & 6  \\
  &     & \textit{color}            & \textit{progressive} & 6  \\
  & ART & \textit{regular}          & \textit{symmetric}   & 8  \\
  &     & \textit{shape}            & \textit{color}       & 8  \\
  & ARC & \textit{crystallographic} & \textit{progressive} & 10 \\
  &     & \textit{shape}            & \textit{isolation}   & 10
\end{tabular}}
\end{table}

\begin{table}[t]
\caption{Precision, Recall and F1 scores of each class of each domain. Average values are given in the last row of the table. }\label{tbl7}
\centering
\resizebox{.48\textwidth}{!}{\begin{tabular}{llllllllll}
 &
  \multicolumn{3}{c}{\textbf{Photography (PHT)}} &
  \multicolumn{3}{c}{\textbf{Art (ART)}} &
  \multicolumn{3}{c}{\textbf{Architecture (ARC)}} \\ \cline{2-10} 
 &
  \multicolumn{1}{c}{Precision} &
  \multicolumn{1}{c}{Recall} &
  \multicolumn{1}{c}{F1} &
  \multicolumn{1}{c}{Precision} &
  \multicolumn{1}{c}{Recall} &
  \multicolumn{1}{c}{F1} &
  \multicolumn{1}{c}{Precision} &
  \multicolumn{1}{c}{Recall} &
  \multicolumn{1}{c}{F1} \\ \cline{2-10} 
\textbf{\textit{asymmetric}}       & 0.66 & 0.74 & 0.70 & 0.62 & 0.60 & 0.61 & 0.42  & 0.48 & 0.45 \\
\textbf{\textit{color}}            & 0.67 & 0.72 & 0.69 & 0.74 & 0.86 & 0.80 & 0.81  & 0.86 & 0.83 \\
\textbf{\textit{crystallographic}} & 0.75 & 0.88 & 0.81 & 0.75 & 0.86 & 0.80 & 0.60  & 0.48 & 0.53 \\
\textbf{\textit{flowing}}          & 0.88 & 0.84 & 0.86 & 0.71 & 0.82 & 0.76 & 0.81  & 0.78 & 0.80 \\
\textbf{\textit{isolation}}        & 0.79 & 0.62 & 0.70 & 0.74 & 0.52 & 0.61 & 0.54  & 0.38 & 0.45 \\
\textbf{\textit{progressive}}      & 0.65 & 0.70 & 0.67 & 0.69 & 0.66 & 0.67 & 0.47  & 0.54 & 0.50 \\
\textbf{\textit{regular}}          & 0.82 & 0.84 & 0.83 & 0.78 & 0.76 & 0.77 & 0.59  & 0.72 & 0.65 \\
\textbf{\textit{shape}}            & 0.72 & 0.52 & 0.60 & 0.59 & 0.48 & 0.53 & 0.46  & 0.48 & 0.47 \\
\textbf{\textit{symmetric}}        & 0.77 & 0.82 & 0.80 & 0.75 & 0.84 & 0.79 & 0.72  & 0.68 & 0.70 \\ 
\textbf{Average}          & 0.75 & 0.74 & 0.74 & 0.71 & 0.71 & 0.70 & 0.60  & 0.60 & 0.60
\end{tabular}}
\end{table}

Precision, Recall and F1 scores of each class of each domain are also calculated for the comparison of test results, as given in Table \ref{tbl7}. Precision is a metric of how many predicted items are relevant: it is the ratio of correct predictions over all the predicted labels. Recall is the metric for a class to measure how many of them are predicted correctly. F1 score is the harmonic mean of precision and recall, thus providing an overall metric to evaluate the success of a prediction model. Overall,  higher average values of precision, recall and F1 are obtained for the photography (PHT) domain. This supports our conjecture that the higher accuracy rates are associated with the number of images and purity of expression in PHT domain. Furthermore, \textit{flowing}, \textit{regular} and \textit{crystallographic} classes get higher values in PHT, while \textit{color} and \textit{symmetric} do so in ART and ARC. It confirms that repetition, variation, and continuity represented in visual compositions in PHT are the best recognised patterns. On the other hand, contrasting colors and symmetry are often used and clearly identified in paintings and buildings, which makes them easily detected by the CNN model for ART and ARC domains. \ 

\subsection{Qualitative Results}

\begin{figure*}
	\centering
		\includegraphics[width=0.8\textwidth,height=\textheight,keepaspectratio]{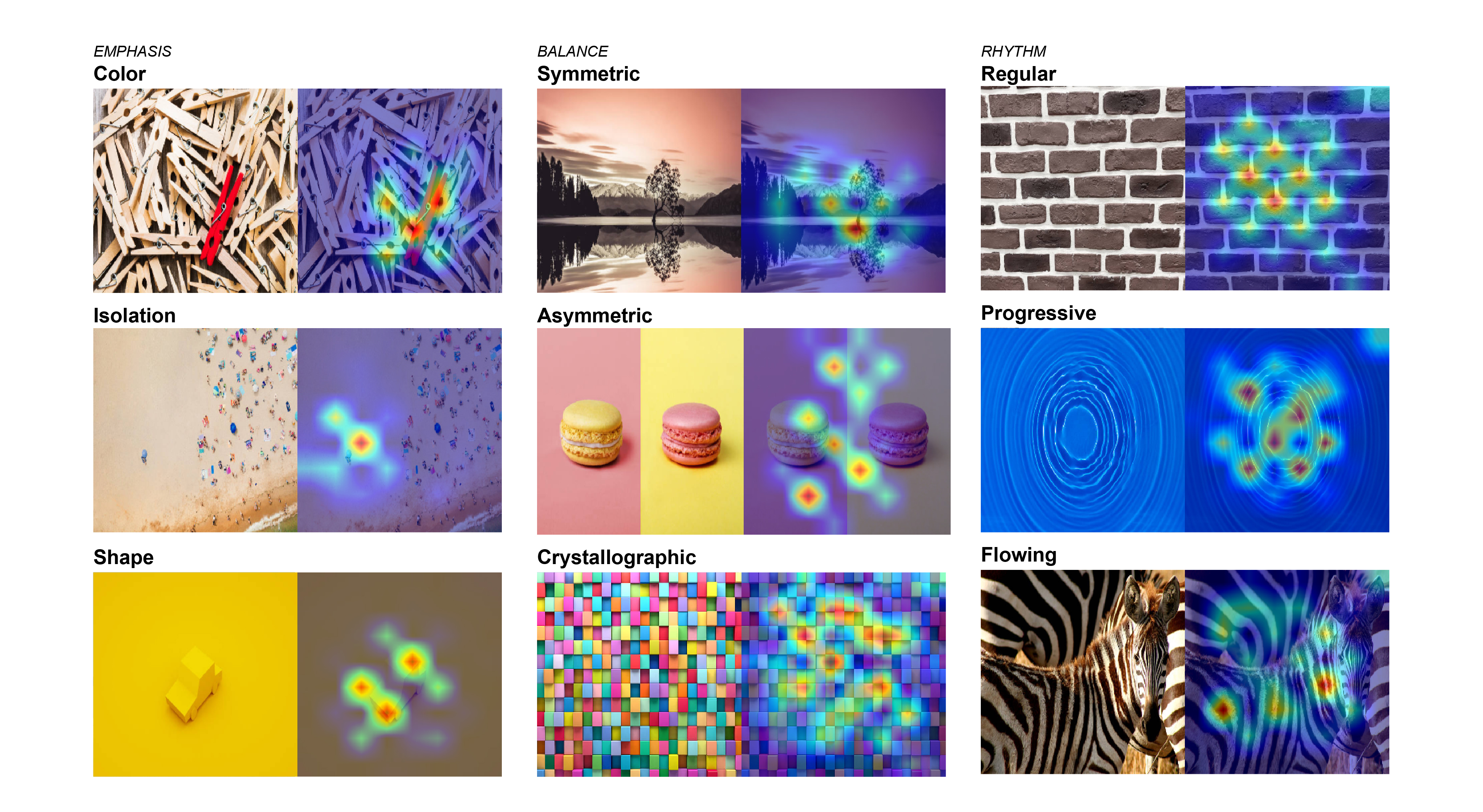}
	\caption{Heatmaps are shown for each sub-VDP. Left and right samples show the original images, and the heatmaps respectively.
    iStock by Getty Images, photography credits (left to right then, top to bottom): /Eblis; /Kanawa Studio; /Miki1988; /Orbon
    Alija; /Ayman-Alakhras; /Goja1; /\begin{otherlanguage}{russian}Дмитрий Ларичев\end{otherlanguage}; /HomePixel; /GezBrowning.}
	\label{F15}
\end{figure*}

The results of the CNN model could be further discussed through an analysis of the heatmaps over images. In order to assess whether the trained CNN model sees meaningful attributes in the images, we set up a heatmap creator using the Grad-CAM method \cite{selvaraju2017grad}. This method projects the output classification spatially back onto the input image space in order to visualize which parts or regions of the input image are having a major effect on the classification output. Some sample heatmaps for corresponding classification results are depicted in Figure \ref{F15} for further analysis. \

On visualizing the important regions over those heatmaps, it can be observed that the network does indeed pay attention to the desired aspects of the visual composition, and does not memorize various unrelated patterns on the images. Accordingly, the locations of the emphasized elements in the images of \textit{color}, \textit{isolation}, and \textit{shape} are visible and focal, which is dependent on the internal local relationships of the design. In contrast, in \textit{crystallographic}, the stressed regions are more spread out, meaning the model makes a correct evaluation on the given principle that affects the design as a whole; hence, rightly, the model focuses on global features for the selected main principle: ‘balance’. Similar observations are made for \textit{regular} and \textit{progressive} where the model highlights the extensive uniformity, or the dynamism in the pattern, respectively. For \textit{symmetric} and \textit{asymmetric}, the magnitude clustered over horizontal and vertical axes, which divide the compositions into two visually equal parts, either by mirroring or value exchanging. Finally, the \textit{flowing} sub-VDP is also accentuated by the model over the curving design elements. These heatmaps provide a qualitative evidence to the computer’s learning to pay attention to relevant global and/or local visual features as needed.  \

\subsection{Human-Computer Competition}
\begin{figure}
	\centering
		\includegraphics[scale=.3]{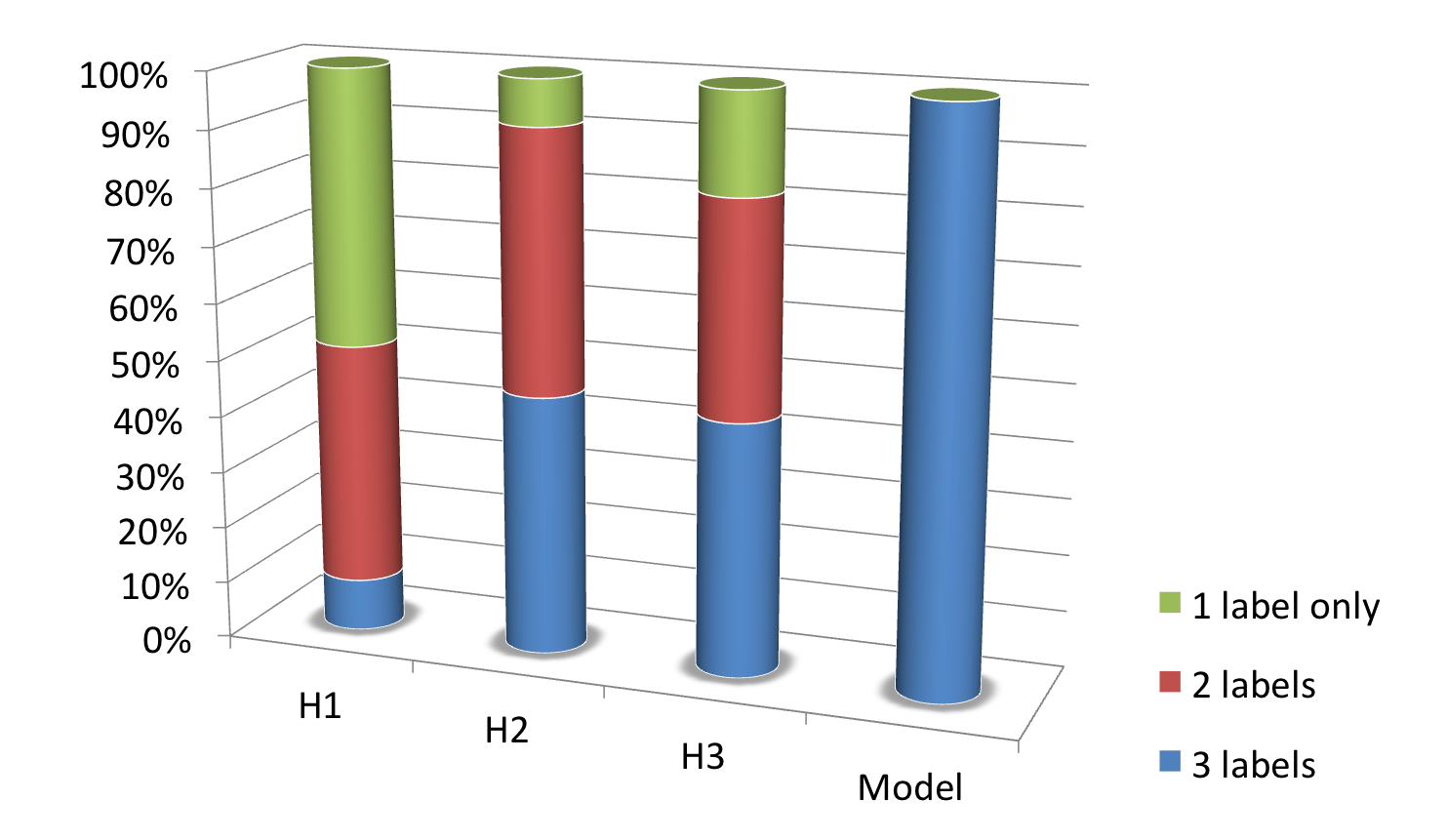}
	\caption{Showing annotation percentages of 1 label, 2 labels and 3 labels given by human experts for 918 images. The Model used “None” never.}
	\label{F17}
\end{figure}

Though the ML model manages success in different experiments, an independent external test that compares AI capability to human experts in interpreting visual design principles is designed. The primary goal of a competition is to observe machine intelligence performance against natural intelligence displayed by humans. We assessed the model by testing its predictions and the value it can generate when applied to a real-world problem related to art and architecture visuals. To that end, we collected a brand new dataset, comprised of approximately 34 images for each of the nine sub-VDP classes in each of the three domains (a total of 918 images). This time, we allowed a multi-label classification, assigning an instance to three labels but in a ranked fashion. It would help us prove the decrease in model performance for some sub-VDPs when categorizing instances to precisely one class. Three professionals (one graphic designer, one academician-artist, and one academician-architect) annotated the new test data using application servers. Human experts are identified as Human 1 (H1), Human 2 (H2), and Human 3 (H3), respectively, in the experiments. They first chose the most relevant class and then additional VDPs for the second and third calls. They could choose “None” if any principle did not match properly with the sample. The labeling distribution is given in Figure \ref{F17}, showing that H1 used single labels for many of the images, whereas H2 and H3 fundamentally filled all three ranks at each instance. Three other independent researchers, including the second author, also labeled the data through a consensus to represent the ground truth or the “oracle.” Our ‘model 1’ in Table \ref{tbl4} is employed as the “Model” in this human-computer competition experiment set-up.  \
 
We first aimed to assess the reliability of agreement between raters by using Fleiss’ kappa. Fleiss’ kappa is a metric used to measure agreement between raters  \cite{fleiss71}. Fleiss’ kappa K is calculated as 0.36 and was measured for the first rank label only, which shows a fair agreement between three raters. The subjective nature of the problem related to the visual perception and aesthetics does not make possible a near-perfect agreement (K= +1) for this situation. Since one of our aims was to assess how effectively the model can demonstrate human performance, we built a distinct scenario by replacing the raters with the model one by one and then calculated Fleiss’ kappa K accordingly. The results are; K= 0.32 (Model instead of Human 1), 0.42 (Model instead of Human 2), and 0.35 (Model instead of Human 3). These results show that the machine learning algorithms can simulate the human expert decision mechanism. 

Building upon the statistical measure of inter-rater reliability, we further checked the matching rate among H1, H2, H3, and the Model. Table \ref{tbl8_1} indicates both the rates of first, second, and third rank label match for each instance and the rate of any single label match per sample without any rank order. It is seen in Table \ref{tbl8_1} that H1–H3 selected the same sub-VDP at the first rank for half the total dataset (0.50). The rates decline for H2–H3 (0.41) and then for H1–H2 (0.39). However, the counts of any single match among all labels of an instance reaches a value of 0.85 using human experts. The model, when again substituted, imitates human reasoning well by exhibiting higher ratings including, 0.53 (H1–Model) for first rank label match and 0.94 (H3–Model) for all order-free label match. To summarize, the AI model develops intelligent behaviors equivalent to that of a human expert in pair-wise experiments.  

\begin{table}[t]
\caption{Matching rates of H1, H2, H3, and the Model; (a) Match for label, take ranking order (b) single match for all labels, ALL DATA.}\label{tbl8_1}
\resizebox{.4\textwidth}{!}{\begin{tabular}{lll}
\textbf{} & \textbf{Ranked Labels} & \textbf{Any Single Label} \\ \hline

Human 1 - Human 2   & 0.3893 (1.)   &  \\
                    & 0.1808 (2.)   & 0.7582 \\
                    & 0.0222 (3.)   &  \\ \hline
Human 2 - Human 3   & 0.4083   &  \\
                    & 0.2511   & 0.8242 \\
                    & 0.2714   &  \\ \hline
Human 1 - Human 3   & 0.4973   &  \\
                    & 0.2038   & 0.8548 \\
                    & 0.1475   &  \\ \hline
Human 1 - Model     & 0.5300   &  \\
                    & 0.2147   & 0.8911 \\
                    & 0.1667   &  \\ \hline
Human 2 - Model     & 0.3595   &  \\
                    & 0.1682   & 0.8900 \\
                    & 0.1643   &  \\ \hline
Human 3 - Model     & 0.4389   &  \\
                    & 0.2000   & 0.9367 \\
                    & 0.1443   &  \\ \hline

\end{tabular}}
\end{table}

We finally compared the accuracy of raters and the learning model against the oracle information, which is available in the form of ground truth. While the oracle does not form a gold standard in this study due to the semantic and conceptual diversity in aesthetic quality, it encodes the input data with corresponding desired output labels \cite{liem2020oracle}. Table \ref{tbl9} lists the accuracy of each rank label. H1 is closest to the oracle first tags with 0.78. Model with 0.55 follows second, where H3 is third (0.54), and H2 is last (0.43). The second and third rank label matches decrease dramatically from the first. We believe that this is expected due to a more fluid existence of multiplicity of sub-VDP in the compositions, while at least a single primary sub-VDP is to be found by construction of the data collection. Accuracy related to any single same label existence with the oracle labels without any rank order reports high conformity. The model represents 0.96, the best performance after H1 (0.97). Domain-specific results also show incredible promise: 0.96 for PHT, 0.96 for ART, and 0.97 for ARC. In short, we conclude that the computer has learned the design principle inherent in visual art and architecture pieces.  

\begin{table}[]
\caption{Accuracy metrics of H1, H2, H3, and the model against the oracle.}\label{tbl9}
\scalebox{0.8}{\begin{tabular}{llllll}
\textbf{Raters} & \textbf{Ranked Labels} & \multicolumn{4}{l}{Any Single Label without Rank Order}        \\
                &                        & \textbf{All Data} & \textbf{PHT} & \textbf{ART} & \textbf{ARC} \\ \hline
Human 1         & 0.7797 (1.)            &                   &              &              &              \\
                & 0.2976 (2.)            & 0.9695            & 0.9658       & 0.9690       & 0.9736       \\
                & 0.2292 (3.)            &                   &              &              &              \\ \hline
Human 2         & 0.4314                 &                   &              &              &              \\
                & 0.1760                 & 0.8192            & 0.8630       & 0.7430       & 0.8581       \\
                & 0.2011                 &                   &              &              &              \\ \hline
Human 3         & 0.5382                 &                   &              &              &              \\
                & 0.2040                 & 0.9159            & 0.9278       & 0.9130       & 0.9076       \\
                & 0.1829                 &                   &              &              &              \\ \hline
Model           & 0.5512                 &                   &              &              &              \\
                & 0.2371                 & 0.9630            & 0.9555       & 0.9628       & 0.9703       \\
                & 0.1706                 &                   &              &              &              \\ \hline
\end{tabular}}
\end{table}

\section{Conclusions}

Our work examines photographs, art visuals and the views of buildings from a visual-aesthetic approach and quantifies basic design principles in their compositions. Conventionally, both for artists and architects, to create a visual composition depends on the usage of the sub-VDP in an original way, and cannot be a prescriptive process. Following this notion, the analytical decomposition of the VDP (with AI) has not been comprehensively explored and inspected in detail until now. We adopt a genuine approach for the solution of this problem by using deep neural networks. For the learning-based model, the main and only input is the cumulative knowledge extracted from a huge collection of carefully-curated products in art and architecture, and their annotations by expert designers. \

Some aspects of our datasets and annotation techniques have some limitations. The amount of data collected for the rules given in Figure \ref{F3} can be equalized, thereby the homogeneity of the data can be improved. The curated datasets naturally contain a level of noise, as they are biased. Employing an extensive public voting mechanism could help overcome that noise to a degree. That would also eliminate the data samples with apparently co-existing sub-VDP in their visual compositions. We also would like to explore our problem as a multi-label classification problem for all datasets, which presents a more inclusive approach when ‘dominant and supporting’ VDP are jointly taken into consideration. Furthermore, the number of labels, as well as the number of the data, can be increased in the future by adding other VDP, such as unity and harmony. \

There can be domain-specific improvements. In the ARC domain, the data is evaluated by considering only the two-dimensional displays of the facade compositions instead of the 3D features. Also, the perspective views of the facades are included in the dataset next to the front, back, right, or left views. These decisions lead to an increasing amount of noise and non-standardization in the architecture dataset, and need to be considered in the future. Our research can also easily be extend to the recognition of the VDP in new design domains such as interior design and industrial design. It can also be improved by domain recognition and learning across given domains based on the VDP through transfer learning \cite{storkey2009training}. \

We can conclude that our ML model detects VDP, and it can provide design support to designers by creating an objective base in the visual aesthetic analysis of any design product, such as the existing buildings in the educational and professional fields. Also, the output of this model can be used with parametric design tools to support a real-time computational design process and can be integrated into the existing design generation methods for a proper visual evaluation of the project proposals. \

\section*{Acknowledgement}

We are grateful to the Museum of Contemporary Art Chicago for
providing copyrights of the artworks, and following photographers and architects for granting
permission to use their photographs: Aitor Ortiz, Tord-Rikard Söderström, Katja
L.(Instagram/@architecturetourist), Altug Galip (Instagram/@kyrenian), Florent Michel, René
de Wit, Tomás Alvarez Robledo, John Gollings, Ieva Saudargaité, Miguel Souto,
SOMOS.Arquitectos, and Peter Kuczia. We are thankful to Cemil Cahit Yavuz, the artist and
his team for data labeling. We wish to thank Murat Can Kurşun, the graphic designer; Prof.Dr.
Yüksel Demir, the architect and; Assoc.Prof.Dr. Oğuz Haşlakoğlu, the artist who put their
expert knowledge and annotated the new dataset in the human vs. computer competition
experiment. We also wish to thank Gülçin Baykal, Dilara Gökçe and Ahmed Hancıoğlu for
their great contributions in data curation. This project was financially supported by Istanbul
Technical University Scientific Research Projects Unit.

\printcredits

\bibliographystyle{unsorted}

\bibliography{cas-refs}


\end{document}